\documentclass[journal]{IEEEtran}
\usepackage{amsmath,amssymb}
\usepackage{mathtools}
\usepackage{xspace}
\usepackage{booktabs}
\usepackage{xcolor}
\usepackage{colortbl}
\usepackage{multirow}
\usepackage{array}
\usepackage{hyperref}
\usepackage{url}
\usepackage{makecell}
\usepackage{changepage}
\usepackage{changes}

\usepackage{framed}
\usepackage{float}
\usepackage{booktabs}
\usepackage{longtable}
\usepackage{nicefrac}
\usepackage{amssymb}
\usepackage{amsmath}
\usepackage{xspace}
\usepackage{mathtools}
\usepackage{diagbox}
\usepackage{graphics}
\usepackage{rotating}
\usepackage[algo2e,ruled,vlined,linesnumbered]{algorithm2e}
\newcommand{\assign}{\leftarrow}
\usepackage{graphicx}
\usepackage{subfigure}
\usepackage{ifthen}
\usepackage{wrapfig}
\usepackage{balance} 
\usepackage{epsfig}
\usepackage{xcolor, colortbl}
\usepackage[normalem]{ulem}
\usepackage{dsfont}

\newcommand{\oea}{$(1 + 1)$~EA\xspace}

\newcommand{\mlga}{$(\mu+\lambda)$~GA\xspace}

\newcommand{\om}{\textsc{OneMax}\xspace}
\newcommand{\onemax}{\om}
\newcommand{\lo}{\textsc{Leading\-Ones}\xspace}

\newcommand{\leadingones}{\lo}

\newcommand{\R}{\mathbb{R}}
\newcommand{\N}{\mathbb{N}}

\newtheorem{definition}{Definition}[section]

\usepackage{graphicx}
\usepackage[utf8]{inputenc}

\title{Automated Configuration of Genetic Algorithms by Tuning for Anytime Performance}

\author{Furong Ye,
        Carola Doerr, 
         Hao Wang, 
   and Thomas B\"ack,~\IEEEmembership{Fellow, IEEE}
   
\IEEEcompsocitemizethanks{\IEEEcompsocthanksitem 
Furong Ye (E-mail: f.ye@liacs.leidenuniv.nl), 
Hao Wang (E-Mail: h.wang@liacs.leidenuniv.nl), and 
Thomas B\"ack (E-Mail: t.h.w.baeck@liacs.leidenuniv.nl) 
are with Leiden Institute of Advanced Computer Science, Leiden University, Leiden,
Netherlands.%
}
\IEEEcompsocitemizethanks{\IEEEcompsocthanksitem Carola Doerr is with Sorbonne Universit\'e, CNRS, LIP6, Paris, France. E-mail: Carola.Doerr@lip6.fr.
}
}

\begin{document}

\maketitle

\begin{abstract}
Finding the best configuration of algorithms' hyperparameters for a given optimization problem is an important task in evolutionary computation. We compare in this work the results of four different hyperparameter optimization approaches for a family of genetic algorithms on 25 diverse pseudo-Boolean optimization problems. More precisely, we compare previously obtained results from a grid search with those obtained from  three automated configuration techniques: iterated racing, mixed-integer parallel efficient global optimization, and mixed-integer evolutionary strategies. 

Using two different cost metrics, expected running time and the area under the empirical cumulative distribution function curve, we find that in several cases the best configurations with respect to expected running time are obtained when using the area under the empirical cumulative distribution function curve as the cost metric during the configuration process. Our results suggest that even when interested in expected running time performance, it might be preferable to use anytime performance measures for the configuration task. We also observe that tuning for expected running time is much more sensitive with respect to the budget that is allocated to the target algorithms. 
\end{abstract}


\section{Introduction}
Mutation and crossover are two key operators of genetic algorithms (GAs), and there is a long debate about the effectiveness of the two operators and their combinations~\cite{MitchellHF93,spears1993crossover}.

\emph{Mutation} can maintain the population diversity of the GA, and it allows the GA to exploit promising search areas by using small mutation rates. 
In the early years of evolutionary computation, \emph{evolution strategies}~\cite{rechenberg1989evolution} concentrated on mutation and showed their power with adaptive mutation rates for optimization~\cite{back1991survey}.

Classic GAs, on the other hand, used \emph{crossover} as their main variation operator~\cite{baeck96,DeJong75,Goldberg89}. Intuitively speaking, the key idea of crossover is to create offspring by recombining information from two or more ``parents''. As is the case for mutation operators~\cite{doerr2017fast,ye2019interpolating},  novel types of crossover keep being developed
~\cite{thierens2011optimal,TinosWCO21,TintosWC15}. 

Several works study the synergy of mutation and crossover, both by empirical and by theoretical means, see~\cite{DeJong75,ELSAYED20111877,Yoon22,Murata96} and~\cite{Sudholt2020chapterdiversity}, respectively, as well as references mentioned therein.
However, most of these results focus on specific algorithms and problems. Widely accepted guidelines for their deployment are sparse, leading to a situation in which users often rely on their own experience. To reduce the bias inherent to such manual decisions, a number of \emph{automated algorithm configuration} techniques have been developed, to assist the user with data-driven suggestions. 
To deploy these techniques, one formulates the operator choice and/or their intensity as a meta-optimization problem, widely referred to as the \emph{algorithm configuration (AC)} or the \emph{hyperparameter optimization (HPO)} problem. 

The AC problem was classically addressed by standard search heuristics such as mixed-integer evolution strategies~\cite{back1994parallel,grefenstette1986optimization,li2013mixed}. More specific AC tools have been developed in recent years, among them surrogate-based models (e.g., SPOT~\cite{bartz2005sequential}, SMAC~\cite{SMAC}, MIP-EGO~\cite{van2019automatic}), racing-based methods (Irace~\cite{lopez2016Irace}, F-race~\cite{birattari2010f}) and optimization-based methods (ParamILS~\cite{ParamILS}).

\textbf{Summary of Our Contributions.} 
In this work, we study the effects of automated algorithm configuration on a genetic algorithm (GA) framework, applied to 25 diverse pseudo-Boolean optimization problems. We compare the results of four different configuration techniques. 
Our main interest is in analyzing the influence of the cost metric that is used to score different configurations on the quality of the configuration suggested by the AC methods.
We consider two different cost metrics: \emph{minimizing the expected running time (ERT)} and \emph{maximizing the area under the empirical cumulative distribution function curve  of running times (AUC)}. 
While minimizing ERT favors the average first hitting time of a single \emph{fixed-target}, maximizing the AUC metric aims at optimizing \emph{anytime performance}, which is measured across a whole set of (budget, target value) pairs. 
We show that in several cases tuning for AUC yields configurations that have smaller ERT values than those that were obtained when directly tuning for ERT.   

Concretely, we build on our previous work~\cite{Yeppsn2020} in which we analyzed a configurable framework of $(\mu+\lambda)$~GAs that scales the relevance of mutation and crossover by means of the crossover probability $p_c \in [0,1]$. The framework creates new solution candidates by applying either mutation (with probability $1-p_c$) or crossover (with probability $p_c$). This way, it can separate the influence of these operators from each other. 
While we have studied several operator choices in~\cite{Yeppsn2020} via plain grid search, we consider here only one type of crossover (uniform crossover) and one type of mutation (standard bit mutation), to keep the search space manageable and to better highlight our key findings. 

We compare in this work the results from~\cite{Yeppsn2020} with those obtained from three different types of automated algorithm configuration methods, one based on iterated racing (we use Irace~\cite{lopez2016Irace}), one surrogate-assisted technique (we use the mixed-integer parallel efficient global optimization MIP-EGO~\cite{van2019automatic}), and a classic heuristic optimization method (we use the mixed-integer evolutionary strategies MIES suggested in~\cite{li2013mixed}). 
Our testbed are the 25 functions from the pseudo-Boolean optimization (PBO) test suite, suggested in~\cite{doerr2019benchmarking,Yeppsn2020} and available in the IOHprofiler benchmarking environment~\cite{IOHprofiler}.

Compared to the \oea and to the configurations obtained by the grid search in~\cite{Yeppsn2020}, we observe that $(1+\lambda)$~mutation-only GAs perform well on \onemax, \leadingones, and most of their so-called W-model extensions (see~Sec. 3.7 in~\cite{doerr2019benchmarking}), and on Ising-Models, whereas the \mlga benefits from using crossover and different mutation rates on the more complex optimization problems. Thanks to its ability to handle conditional configuration spaces, Irace is the only method that finds configurations of $(1+\lambda)$~mutation-only GAs for problems on which these are efficient.
We also observe that, on some problems, the automatic configurators cannot obtain hyperparameter settings that are as good as those provided by a simple grid search.
However, our key finding is that the configuration methods can find better configurations in terms of the ERT by using AUC as the cost metric, compared to directly using ERT instead. This advantage is particularly pronounced when the budget of the GAs is small compared to the ERT obtained by the best possible configuration. In this case, ERT cannot reflect the differences among the configurations, and the anytime performance provides more information to guide the configuration process. 

\textbf{Related Work.}  
Automated algorithm configuration for improving the anytime performance of algorithms has been applied in several works, both with respect to classical CPU time (e.g., for the traveling salesman problem~\cite{bossek2020anytime}, for a MAX-MIN ant system~\cite{lopez2014automatically}, and for mixed-integer programming ibid) and for the here-considered function evaluation budgets (see~\cite{AmineGECCO} for a recent example). 
Sensitivity of algorithm configuration approaches with respect to the cutoff time (i.e., the budget of function evaluations for which algorithms are evaluated) has been studied in~\cite{HallOS22impact,hall2020analysis}. In particular, \cite{hall2020analysis} provides an example in which tuning for fixed-budget performance can be preferable under small cutoff times than tuning for optimization time, even if the latter is the original objective.
We are not aware of any works using anytime performance measures with the objective to identify configurations that minimize ERT values.

\textbf{Availability of Data and Source Code.}
All our data is publicly available at~\cite{Dataref} and at the public repository of IOHanalyzer, accessible via \url{https://iohanalyzer.liacs.nl/}. Users interested in an interactive evaluation of our data can find our data by selecting `PBO' as dataset source and `2021-mlga' as dataset. After loading the data from the repository, various statistics and visualizations are available for further investigation,  see~\cite{IOHanalyzer} for an overview. 
The source code of our experiments is available at  \url{https://github.com/FurongYe/Configuration-of-Genetic-Algorithms}.

For readers' convenience, we provide in the supplementary material fixed-target charts for each of the 25 functions, similar to the one presented in Fig.~\ref{fig:F10}.

\textbf{Paper Organization.} 
Sec.~\ref{sec:setup} describes the configurable \mlga, the benchmark problems, and the cost metrics used to evaluate the algorithms. In Sec.~\ref{sec:conf} we present results for the three selected AC methods, evaluate their performance against the \oea, and discuss the impact of the cost metrics for the AC methods. 
We conclude our work and discuss directions for future research in Sec.~\ref{sec:conclude}.

\section{Preliminaries}
\label{sec:setup}
In this section, we describe our \mlga framework (Sec.~\ref{sec:alogs}), the benchmark problems (Sec.~\ref{sec:problems}), and the performance measures used to evaluate the algorithms (Sec.~\ref{sec:performanmeasure}). 

\textbf{Notational conventions:} Throughout this paper, we study the maximization of so-called pseudo-Boolean functions, i.e., functions of the type $f:\{0,1\}^n \to \R$, mapping length-$n$ bit strings to real numbers. In particular, we always denote by $n$ the dimension of the problem. For each positive integer $k$, we abbreviate $[k]=\{j \in \N \mid 1 \le j \le k\}$ and $[0..k]=[k] \cup \{0\}$. 

\subsection{A Family of $(\mu+\lambda)$~Genetic Algorithms}
\label{sec:alogs}

To study how mutation, crossover, and their combinations can be beneficial for GAs, we investigate a configurable genetic algorithm framework that can switch from mutation-only GAs towards crossover-only GAs by tuning the crossover probability $p_c$. 
Details of the framework can be found in Alg.~\ref{alg:GA}, and we refer to it as the \mlga in the following.

The \mlga initializes the population uniformly at random (u.a.r.) and terminates when the optimum is found or when the cutoff time (i.e., the maximal number of fitness evaluations that the algorithm may perform) is reached.
For each iteration, the \mlga creates $\lambda$ offspring. Each offspring is created by applying \emph{either} mutation \emph{or} crossover. 
More precisely, crossover is applied with probability $p_c$ and mutation is used otherwise.  
Consequently, the \mlga with $p_c = 0$ refers to a mutation-only GA, and the \mlga with $p_c = 1$ refers to a crossover-only GA. 
After each iteration, the best $\mu$ points out of the combined set of $\mu$ parents and $\lambda$ offspring form the next set of parents (``plus selection''). 

For this work, we consider \emph{uniform crossover} and \emph{standard bit mutation} with mutation rate $p_m$. 
These two variation operators are briefly described as follows. Note that the framework can easily be extended to include other operators.
\begin{itemize}
    \item \emph{Uniform crossover}: Each bit of the offspring is copied either from the first or from second parent, chosen independently and u.a.r. for each position.  
    \item \emph{Standard bit mutation}: The offspring is created by first copying the parent and then flipping $\ell$ distinct bits that are chosen u.a.r. The value $\ell$ is called the \emph{mutation strength}. It is sampled from a conditional binomial distribution Bin$_{>_0}(n,p_m)$, which assigns to each $k \in [n]$ a probability of $\binom{n}{k}p_m^k(1-p_m)^{n-k}/(1-(1-p_m)^n)$. Compared to the classic textbook description of standard bit mutation, this version avoids sampling copies of the parents, by allocating the probability of sampling $\ell=0$ proportionally to all positive values $\ell \in [n]$; see~\cite{PintoD18PPSN} for a discussion of this ``resampling'' strategy.
    \end{itemize}

\begin{algorithm2e}[t]
\textbf{Input:} Population sizes $\mu$, $\lambda$, crossover probability $p_c$, mutation rate $p_m$, cutoff time $B$\;
\textbf{Initialization:} 
    \For{$i=1,\ldots,\mu$}{
    Sample $x^{(i)} \in \{0,1\}^n$ u.a.r.\;
    Evaluate $f(x^{(i)})$
    }
		Set $P = \{x^{(1)},x^{(2)},..., x^{(\mu)}\}$\;
	\textbf{Optimization:}
	\For{$t=1,2,3,\ldots$}{
	    $P' \leftarrow \emptyset$\;
	    \For{$i=1,\ldots,\lambda$}{
	        Sample $r \in [0,1]$ u.a.r.\;
	        \eIf{$r \le p_c$}{
	            Select two individuals $x,y \in P$ u.a.r. (with replacement)\;
	            $z^{(i)} \assign \text{Crossover}(x,y)$\;
	            \lIf{$z^{(i)} \notin \{x,y\}$}{evaluate $f(z^{(i)})$ \\
	            \textbf{else} Infer $f(z^{(i)})$ from parent}
	        }{
	            Select an individual $x \in P$ u.a.r.\;
	            $z^{(i)} \assign \text{Mutation}(x)$\;
	            \lIf{$z^{(i)} \neq x$}{evaluate $f(z^{(i)})$ 
	            \\ \textbf{else} Infer $f(z^{(i)})$ from parent}
	        }
	        $P'\leftarrow P'\cup\{z^{(i)}\}$\;
	    }
        $P$ is updated by the best $\mu$ points in $P \cup P'$ (ties broken u.a.r.)\;
        \textbf{Terminate} the algorithm if the optimum is found or if the number of evaluations (at line 13 and line 18) exceeds the given cutoff time.
	}
\caption{A Family of $(\mu+\lambda)$~Genetic Algorithms}
\label{alg:GA}
\end{algorithm2e}


\subsection{The IOHprofiler Problem Suite PBO}
\label{sec:problems}
The PBO problem suite suggested in~\cite{doerr2019benchmarking,Yeppsn2020} and available in the IOHprofiler benchmarking environment~\cite{IOHprofiler} contains $25$ pseudo-Boolean optimization problems. The definitions of these problems is available in the supplementary material. We provide here  only a short summary:
\textbf{F1} is \onemax, \textbf{F2} is \leadingones, \textbf{F3} is a linear function with harmonic weights, \textbf{F4-10} and \textbf{F11-17} are so-called W-model extensions of F1 and F2, respectively,\footnote{The W-model transformations were suggested in~\cite{weise2018difficult,WmodelInstancesASoC} and later extended in~\cite{doerr2019benchmarking}. In a nutshell, these transformations introduce dummy variables that do not impact the quality of a solution (``dummy'' transformation), perturb the fitness values (``ruggedness''), reduce the string size by partitioning the bit string and applying a majority vote within each substring (``neutrality''), or map substrings to different ones (``epistasis'').}
\textbf{F18} is the Low Autocorrelation Binary Sequences (LABS) problem, \textbf{F19-21} are Ising Model problems, \textbf{F22} is an instance of the Maximum Independent Vertex Set (MIVS) problem, \textbf{F23} is the N-Queens problem (NQP), \textbf{F24} is a Concatenated Trap (CT) problem, and \textbf{F25} is a randomly chosen NK Landscape instance. 

Our algorithms are unbiased in the sense of Lehre and Witt~\cite{LehreW12}, i.e., they are invariant under reordering of the bit string and under XOR of the solution candidates with arbitrary strings. They are also comparison-based, making them invariant with respect to strictly motononic transformations of the function values. We can therefore restrict our experiments to a single problem instance, and do not make use of IOHprofiler's option to transform the instances into problems with isomorphic fitness landscape and/or translated fitness values. More precisely, we tune on the first instance of each problem in the PBO suite.

\subsection{Algorithm Performance Measures}
\label{sec:performanmeasure}
As mentioned in the introduction, we use two main performance criteria: \emph{expected running time (ERT)} and the \emph{area under the curve representing the empirical cumulative distribution function (AUC)}. While the ERT concerns only the hitting time for a single ``final'' target, AUC considers a set of targets. The formal definitions are as follows. 

\begin{definition}[ERT: Expected Running Time] Given a target value $\phi$ for a problem $P$, the ERT of algorithm $A$ for hitting $\phi$ is
\begin{equation}
\text{ERT}(A,P,\phi) = \frac{\sum_{i=1}^{r}\min\{t_i(A,P,\phi),B\}}{\sum_{i=1}^{r}\mathds{1}\{t_i(A,P,\phi)< \infty\}} \;,
\end{equation}
where $r$ is the number of independent runs of $A$, $B$ is the cutoff time (i.e., the maximal number of solution candidates that algorithm $A$ may evaluate in one run), $t_i(A,P,\phi) \in \N \cup \{\infty\}$ is the running time (for finite values, the running time is the number of function evaluations that the $i$-th run of $A$ on problem $P$ uses to hit the target $\phi$ and $t_i(A,P,\phi)=\infty$ is used if none of the solutions is better than $\phi$), and $\mathds{1}(\mathcal{E})$ is the indicator function returning 1 if event $\mathcal{E}$ is true and 0, otherwise. If the algorithm hits the target $\phi$ in all $r$ runs, the ERT is equal to the average hitting time (AHT).
\label{def:ERT}
\end{definition}

\begin{definition}[ECDF: empirical cumulative distribution function of the running time]  Given a set of targets $\Phi = \{\phi_i \in \R \mid i \in [m]\}$  for a real-valued problem $P$ and a set of budgets $T=\{t_j \in [B] \mid j \in [z]\}$ for an algorithm $A$, the ECDF value of $A$ at budget $t_j$ is the fraction of (run, target)-pairs $(s,\phi_i)$ that satisfy that in run $s$ of algorithm $A$ at least one of the first $t_j$ evaluated solutions has fitness at least as good as $\phi_i$.
\end{definition}

\begin{definition}[AUC: area under the ECDF running time curve] Given a set of targets $\Phi = \{\phi_i \in \R \mid i \in [m]\}$ and a set of budgets $T=\{t_j \in [B] \mid j \in [z]\}$, the AUC~$\in [0,1]$ (normalized over $B$) of algorithm $A$ on problem $P$ is the area under the ECDF curve of the running time over multiple targets. For maximization, it reads,
\begin{equation*}
\text{AUC}(A,P,\Phi,T) = \frac{\sum\limits_{h=1}^{r}\sum\limits_{i=1}^{m}\sum\limits_{j=1}^{z} \mathds{1} \{\phi_h(A,P,t_j) \ge \phi_i\}} {r m z} \;,
\label{def:AUC}
\end{equation*}
where $r$ is the number of independent runs of $A$ and $\phi_h(A,P,t)$ denotes the value of the best solution that $A$ evaluated within its first $t$ evaluations of the run $h$.
\end{definition}

\begin{table*}[t]
\small
\setlength{\tabcolsep}{2.5pt}
\centering
\caption{\textnormal{Target values used to compute the ERT metric on each problem.}} 
\label{tab:targets-f}
\begin{tabular}{r|c|c|c|c|c|c|c|c|c|c|c|c|c|c|c|c|c|c|c|c|c|c|c|c|c}
\hline
funcId & $1$ & $2$ & $3$ & $4$ & $5$ & $6$ & $7$ & $8$ & $9$ & $10$ & $11$ & $12$ & $13$ & $14$ & $15$ & $16$ & $17$ & $18$ & $19$ & $20$ & $21$  & $22$ & $23$ & $24$ & $25$\\\hline
Target & 100 & 100 & 5\,050 & 50 & 90 & 33 & 100 & 51 & 100 & 100 & 50 & 90 & 33 & 7 & 51 & 100 & 100 & 4.216 & 98 & 180 & 260 &  42 & 9 & 17.196 & -0.297 \\
\hline
\end{tabular}
\end{table*}

\section{Algorithm Configuration for Benchmarking}
\label{sec:ConfForBench}\label{sec:exp}

\label{sec:conf}
To explore promising configurations of the \mlga, we study its associated algorithm configuration problem: 
\begin{definition}[Algorithm Configuration (AC) Problem] 
Given a problem $P$, a parametrized algorithm $A$ with parameter space $\Theta$, and a cost metric $c:\Theta \times P \rightarrow \mathcal{R}$ that is subject to minimization, the objective of the AC problem is to find a configuration $\theta^* \in \underset{\theta\in\Theta}{\arg \min } \; c(\theta,P)$.
\end{definition}
Note that this definition is adjusted to our setting in which we tune for a single problem instance. It therefore ignores aggregation over multiple problem instances, hence the deviation from other common definitions as, for example, available in~\cite{eggensperger2019pitfalls,Stutzle2019}.

\subsection{Algorithm Configurators Utilized}
We apply three AC methods to configure the \mlga: Irace~\cite{lopez2016Irace}, a mixed-integer parallel efficient global optimization (MIP-EGO~\cite{wang2017new}), and the mixed-integer evolution strategy (MIES~\cite{li2013mixed}), which we briefly describe in the following paragraphs. All configurators work with a user-defined \emph{configuration budget}, which is the maximal number of algorithm runs that the AC method is allowed to perform before recommending its final outcome.  

\textbf{Irace} \cite{lopez2016Irace} is a so-called iterated racing method designed for hyperparameter optimization. 
The main steps of Irace are (1) sampling hyperparameter values from distributions that evolve over time, (2) evaluating the configurations across a set of instances, (3) applying a racing method (i.e., statistical tests) to select the configurations that are not further evaluated in the subsequent rounds, and (4) updating the sampling distributions.
At the end of the configuration process, one or several \emph{elite} configurations are returned to the user, along with their performance observed during the configuration. The distributions from which the hyperparameters are sampled are independent of each other,  unless the user specifies constraints and/or other requirements. 

\textbf{Efficient global optimization (EGO)}, also known as Bayesian optimization, is designed to solve costly-to-evaluate global optimization problems. 
For our configuration problem, we use an EGO-variant called mixed-integer parallel EGO (MIP-EGO~\cite{wang2017new}) 
capable of handling mixed-integer search space.
EGO starts with an initial set of solution candidates $\{\theta_1,\theta_2,...\}$ and evaluating their fitness $\{c_1,c_2,...\}$. From these observations, EGO learns
a predictive distribution of the fitness value for each unseen configuration using stochastic models, e.g., random forests or Gaussian processes. Aiming at balancing the trade-off between the accuracy and uncertainty of this predictive distribution, EGO uses a so-called \emph{acquisition function} to decide which solution candidates to sample next. Common acquisition functions are \emph{expected improvement} and \emph{probability of improvement}; see~\cite{ShahriariSWAF16,abs-1807-02811} for an overview. For this work, we use the moment-generating function of improvement (MGFI)~\cite{wang2017new}, which is defined as the weighted combination of all moments of the predictive distribution. For the weights, we used the setting recommended in~\cite{WangEB18}.
To learn the predictive distribution, we choose a random forest model as it deals with the mixed-integer/categorical search space more naturally than Gaussian processes.

\textbf{The mixed-integer evolution strategy (MIES)} uses principles from evolution strategies for handling continuous, discrete, and nominal parameters by using self-adaptive mutation operators for all three parameter types~\cite{li2013mixed}.
MIES starts with a randomly initialized parent population of size $\mu$. In each iteration it generates $\lambda$ offspring, by recombining two randomly selected parents and then mutating the solution resulting from their recombination, after which $(\mu,\lambda)$ selection is applied to the offspring population, i.e., the $\mu$ best of the $\lambda$ offspring form the parent population of the next iteration. 

\subsection{Experimental Setup} 
Each AC method is granted a budget of $5\,000$ \emph{target runs}, where each target run corresponds to ten independent runs of the \mlga using the configuration that the AC method wishes to evaluate. As previously mentioned, we use ERT and AUC as performance metric, and these values are computed from the ten independent runs.

Irace requires a set of instances for the tuning process. We imitate these 
\emph{instances} by the independent runs of the (stochastic) solvers. This is in line with previous approaches, suggested, for example, in~\cite{DangD19}.
MIP-EGO starts with $10$ initial candidates by the default setting of the package~\cite{wang2017new}. 
We use a $(4,28)$-MIES, following the parameter settings suggested in~\cite{li2013mixed}. 

To obtain a useful baseline against which we can compare other algorithms, we configure the \mlga on each PBO problem separately.
We focus on the problems in dimension $n=100$. 

Once the optimized parameter settings are obtained from the algorithm configuration, we perform 100 \emph{independent validation runs} with the suggested parameter settings. This is to have a fair comparison, as the number of algorithm evaluations may differ quite drastically between the configurators and between different runs of the same configurator. All performance comparisons discussed in this paper are made with respect to these validation runs. 

For evaluating GA candidates during the configuration process, the ERT values are calculated with respect to the targets listed in Tab.~\ref{tab:targets-f} and to the cutoff time of $50\,000$ function evaluations. This is half the budget used in~\cite{Yeppsn2020}, but, according to the results presented there, our cutoff time is still larger than the number of function evaluations needed to hit the corresponding targets -- except for F18, which is a very challenging problem.

For the AUC, the set of targets are $100$ values, equally spaced in the interval $[\phi_{\min},\phi_{\max}]$, where $\phi_{\max}$ is equal to the ERT targets listed in Tab.~\ref{tab:targets-f} and $\phi_{\min}$ is equal to 0 except for the following functions:  $\phi_{\min}=-19\,590$ for function F22, $\phi_{\min}=-3\,950\,000$ for function F23, and  $\phi_{\min}=-1$ for function F25. We evaluate the AUC at each budget, i.e., in the notation of Definition~\ref{def:AUC} we use $T=[50\,000]$.

\begin{table*}[htb]
\centering
\caption{
\textnormal{	
Configurations of the \mlga obtained by grid search, Irace, MIP-EGO, and MIES. C indicates the cost metric used by the configurators. Results for maximizing AUC are  obtained independently from those obtained for minimizing ERT. 
}
}
\label{tab:config}
\begin{tabular}{|r|c||c|c|c|c||c|c|c|c||c|c|c|c||c|c|c|c|}
\hline																																			
 \multirow{2}{*}{\textbf{F}} & \multirow{2}{*}{\textbf{C}} & \multicolumn{4}{c||}{\textbf{Grid Search}}  & \multicolumn{4}{c||}{\textbf{Irace}} & \multicolumn{4}{c||}{\textbf{MIP-EGO}}& \multicolumn{4}{c|}{\textbf{MIES}}\\																																			
  \cline{3-18}																															
 & & $\boldsymbol{\mu}$ &$\boldsymbol{\lambda}$ & $\boldsymbol{p_m}$ & $\boldsymbol{p_c}$ & $\boldsymbol{\mu}$ &$\boldsymbol{\lambda}$ & $\boldsymbol{p_m}$ & $\boldsymbol{p_c}$& $\boldsymbol{\mu}$ &$\boldsymbol{\lambda}$ & $\boldsymbol{p_m}$ & $\boldsymbol{p_c}$ & $\boldsymbol{\mu}$ &$\boldsymbol{\lambda}$ & $\boldsymbol{p_m}$ & $\boldsymbol{p_c}$\\																																			
 \hline																																			
\multirow{2}{*}{1}	&	ERT	&	10	&	1	&	0.01	&	0.5	&	15	&	5	&	0.007	&	0.782	&	3	&	4	&	0.024	&	0.774	&	2	&	2	&	0.006	&	0.778	\\
	&	AUC	&	10	&	1	&	0.01	&	0.5	&	1	&	1	&	0.006	&	0	&	3	&	11	&	0.006	&	0.208	&	2	&	2	&	0.005	&	0.521	\\
 \hline																																			
\multirow{2}{*}{2}	&	ERT	&	10	&	1	&	0.01	&	0.5	&	1	&	39	&	0.006	&	0	&	2	&	9	&	0.006	&	0.008	&	2	&	3	&	0.005	&	0.026	\\
	&	AUC	&	10	&	1	&	0.01	&	0	&	1	&	34	&	0.013	&	0	&	22	&	5	&	0.009	&	0.461	&	2	&	1	&	0.008	&	0.005	\\
 \hline																																			
\multirow{2}{*}{3}	&	ERT	&	10	&	1	&	0.01	&	0.5	&	4	&	32	&	0.006	&	0.867	&	3	&	24	&	0.009	&	0.391	&	2	&	3	&	0.005	&	0.357	\\
	&	AUC	&	10	&	1	&	0.01	&	0.5	&	4	&	23	&	0.011	&	0.550	&	2	&	3	&	0.007	&	0.511	&	2	&	1	&	0.006	&	0.203	\\
 \hline																																			
\multirow{2}{*}{4}	&	ERT	&	10	&	1	&	0.01	&	0.5	&	1	&	1	&	0.013	&	0	&	2	&	3	&	0.022	&	0.655	&	2	&	2	&	0.024	&	0.198	\\
	&	AUC	&	10	&	1	&	0.01	&	0.5	&	1	&	1	&	0.013	&	0	&	12	&	9	&	0.052	&	0.668	&	2	&	1	&	0.055	&	0.614	\\
 \hline																																			
\multirow{2}{*}{5}	&	ERT	&	10	&	5	&	0.01	&	0.5	&	3	&	17	&	0.005	&	0.415	&	4	&	2	&	0.026	&	0.294	&	2	&	3	&	0.006	&	0.321	\\
	&	AUC	&	10	&	1	&	0.01	&	0.5	&	8	&	8	&	0.012	&	0.882	&	11	&	20	&	0.008	&	0.318	&	2	&	2	&	0.006	&	0.358	\\
 \hline																																			
\multirow{2}{*}{6}	&	ERT	&	10	&	10	&	0.01	&	0	&	30	&	23	&	0.104	&	0.849	&	3	&	11	&	0.031	&	0.033	&	24	&	12	&	0.135	&	0.908	\\
	&	AUC	&	10	&	10	&	0.01	&	0.5	&	1	&	1	&	0.033	&	0	&	2	&	1	&	0.032	&	0.424	&	2	&	3	&	0.059	&	0.236	\\
 \hline																																			
\multirow{2}{*}{7}	&	ERT	&	50	&	50	&	0.01	&	0	&	45	&	22	&	0.460	&	0.887	&	95	&	46	&	0.018	&	0.971	&	54	&	18	&	0.268	&	0.907	\\
	&	AUC	&	10	&	10	&	0.01	&	0	&	1	&	58	&	0.028	&	0	&	2	&	20	&	0.016	&	0.082	&	23	&	27	&	0.025	&	0.672	\\
 \hline																																			
\multirow{2}{*}{8}	&	ERT	&	10	&	10	&	0.01	&	0.5	&	25	&	48	&	0.014	&	0.643	&	78	&	7	&	0.087	&	0.991	&	5	&	9	&	0.009	&	0.507	\\
	&	AUC	&	10	&	10	&	0.01	&	0.5	&	3	&	6	&	0.015	&	0.441	&	11	&	12	&	0.007	&	0.499	&	21	&	25	&	0.011	&	0.800	\\
 \hline																																			
\multirow{2}{*}{9}	&	ERT	&	100	&	50	&	0.01	&	0.5	&	64	&	90	&	0.056	&	0.938	&	76	&	10	&	0.488	&	0.977	&	54	&	32	&	0.013	&	0.643	\\
	&	AUC	&	50	&	25	&	0.01	&	0.5	&	46	&	40	&	0.029	&	0.815	&	6	&	7	&	0.020	&	0.158	&	18	&	25	&	0.025	&	0.505	\\
 \hline																																			
\multirow{2}{*}{10}	&	ERT	&	100	&	50	&	0.01	&	0.5	&	95	&	69	&	0.325	&	0.978	&	87	&	84	&	0.478	&	0.962	&	79	&	17	&	0.056	&	0.320	\\
	&	AUC	&	100	&	1	&	0.01	&	0.5	&	99	&	13	&	0.085	&	0.982	&	76	&	16	&	0.321	&	0.986	&	92	&	32	&		0.413	& 0.989	\\
 \hline																																			
\multirow{2}{*}{11}	&	ERT	&	10	&	1	&	0.01	&	0	&	1	&	72	&	0.031	&	0	&	4	&	57	&	0.031	&	0.004	&	2	&	2	&	0.025	&	0.000	\\
	&	AUC	&	10	&	1	&	0.01	&	0	&	1	&	13	&	0.035	&	0	&	2	&	1	&	0.040	&	0.003	&	2	&	1	&	0.042	&	0.016	\\
 \hline																																			
\multirow{2}{*}{12}	&	ERT	&	10	&	1	&	0.01	&	0	&	1	&	32	&	0.006	&	0	&	4	&	1	&	0.012	&	0.016	&	2	&	1	&	0.005	&	0.000	\\
	&	AUC	&	10	&	1	&	0.01	&	0	&	1	&	29	&	0.010	&	0	&	2	&	1	&	0.017	&	0.016	&	2	&	1	&	0.014	&	0.005	\\
 \hline																																			
\multirow{2}{*}{13}	&	ERT	&	10	&	10	&	0.01	&	0	&	1	&	10	&	0.044	&	0	&	10	&	19	&	0.032	&	0.282	&	2	&	5	&	0.032	&	0.008	\\
	&	AUC	&	10	&	10	&	0.01	&	0	&	1	&	1	&	0.044	&	0	&	2	&	4	&	0.038	&	0.023	&	3	&	8	&	0.047	&	0.036	\\
 \hline																																			
\multirow{2}{*}{14}	&	ERT	&	100	&	100	&	0.01	&	0.5	&	1	&	60	&	0.343	&	0	&	57	&	34	&	0.077	&	0.988	&	57	&	23	&	0.028	&	0.012	\\
	&	AUC	&	100	&	50	&	0.01	&	0	&	48	&	83	&	0.409	&	0.812	&	5	&	80	&	0.456	&	0.287	&	54	&	9	&	0.481	&	0.480	\\
 \hline																																			
\multirow{2}{*}{15}	&	ERT	&	10	&	10	&	0.01	&	0	&	5	&	64	&	0.014	&	0.042	&	5	&	15	&	0.008	&	0.018	&	6	&	67	&	0.011	&	0.023	\\
	&	AUC	&	10	&	10	&	0.01	&	0	&	21	&	71	&	0.020	&	0.165	&	3	&	9	&	0.016	&	0.078	&	2	&	3	&	0.011	&	0.024	\\
 \hline																																			
\multirow{2}{*}{16}	&	ERT	&	10	&	10	&	0.01	&	0	&	6	&	34	&	0.011	&	0.025	&	2	&	2	&	0.006	&	0.069	&	2	&	4	&	0.005	&	0.002	\\
	&	AUC	&	10	&	10	&	0.01	&	0	&	1	&	1	&	0.010	&	0.000	&	22	&	66	&	0.009	&	0.138	&	2	&	3	&	0.007	&	0.001	\\
 \hline																																			
\multirow{2}{*}{17}	&	ERT	&	10	&	10	&	0.01	&	0.5	&	60	&	57	&	0.133	&	0.512	&	79	&	62	&	0.345	&	0.430	&	2	&	3	&	0.006	&	0.003	\\
	&	AUC	&	10	&	10	&	0.01	&	0	&	1	&	1	&	0.019	&	0	&	2	&	5	&	0.012	&	0.007	&	2	&	4	&	0.019	&	0.034	\\
 \hline																																			
\multirow{2}{*}{18}	&	ERT	&	100	&	50	&	0.01	&	0	&	50	&	44	&	0.006	&	0.010	&	13	&	77	&	0.019	&	0.002	&	26	&	21	&	0.006	&	0.002	\\
	&	AUC	&	10	&	5	&	0.01	&	0	&	20	&	10	&	0.005	&	0.031	&	2	&	22	&	0.006	&	0.011	&	23	&	83	&	0.006	&	0.010	\\
 \hline																																			
\multirow{2}{*}{19}	&	ERT	&	10	&	10	&	0.01	&	0.5	&	98	&	94	&	0.174	&	0.324	&	2	&	5	&	0.014	&	0.286	&	2	&	12	&	0.006	&	0.061	\\
	&	AUC	&	50	&	50	&	0.01	&	0	&	1	&	5	&	0.005	&	0	&	7	&	18	&	0.014	&	0.834	&	2	&	7	&	0.005	&	0.133	\\
 \hline																																			
\multirow{2}{*}{20}	&	ERT	&	50	&	50	&	0.01	&	0	&	1	&	32	&	0.008	&	0	&	6	&	53	&	0.007	&	0.012	&	5	&	66	&	0.005	&	0.033	\\
	&	AUC	&	10	&	10	&	0.01	&	0	&	1	&	30	&	0.013	&	0	&	5	&	7	&	0.011	&	0.483	&	6	&	75	&	0.006	&	0.095	\\
 \hline																																			
\multirow{2}{*}{21}	&	ERT	&	10	&	5	&	0.01	&	0	&	1	&	86	&	0.014	&	0	&	7	&	52	&	0.009	&	0.144	&	5	&	66	&	0.009	&	0.077	\\
	&	AUC	&	10	&	10	&	0.01	&	0	&	4	&	33	&	0.006	&	0.412	&	7	&	9	&	0.006	&	0.833	&	2	&	3	&	0.005	&	0.219	\\
 \hline																																			
\multirow{2}{*}{22}	&	ERT	&	100	&	100	&	0.01	&	0	&	16	&	48	&	0.025	&	0.783	&	18	&	12	&	0.019	&	0.820	&	2	&	3	&	0.005	&	0.270	\\
	&	AUC	&	10	&	10	&	0.01	&	0.5	&	1	&	1	&	0.020	&	0	&	2	&	3	&	0.007	&	0.250	&	2	&	2	&	0.010	&	0.088	\\
 \hline																																			
\multirow{2}{*}{23}	&	ERT	&	10	&	1	&	0.01	&	0	&	62	&	39	&	0.006	&	0.252	&	10	&	40	&	0.013	&	0.010	&	5	&	1	&	0.006	&	0.006	\\
	&	AUC	&	10	&	5	&	0.01	&	0	&	29	&	25	&	0.005	&	0.016	&	4	&	4	&	0.006	&	0.170	&	8	&	16	&	0.005	&	0.038	\\
 \hline																																			
\multirow{2}{*}{24}	&	ERT	&	100	&	50	&	0.01	&	0	&	53	&	85	&	0.025	&	0.004	&	58	&	45	&	0.031	&	0.077	&	21	&	75	&	0.060	&	0.014	\\
	&	AUC	&	10	&	5	&	0.01	&	0	&	7	&	86	&	0.026	&	0.776	&	8	&	3	&	0.032	&	0.277	&	8	&	33	&	0.045	&	0.273	\\
 \hline																																			
\multirow{2}{*}{25}	&	ERT	&	50	&	1	&	0.01	&	0	&	13	&	52	&	0.296	&	0.170	&	68	&	91	&	0.024	&	0.801	&	98	&	40	&	0.009	&	0.941	\\
	&	AUC	&	10	&	10	&	0.01	&	0	&	26	&	53	&	0.028	&	0.790	&	4	&	3	&	0.023	&	0.107	&	9	&	1	&	0.021	&	0.535	\\
 \hline

\end{tabular}
\end{table*}
\begin{table*}[htb]
\centering
\caption{
\textnormal{Absolute ERT and AUC values for the \oea and relative improvement of ERT and AUC for the configurations suggested by the four configuration methods, in comparison against the \oea values. Compared to the ERT and AUC values of the \oea on each problem (indicated by ``EA''), the relative improvement obtained from the automated configuration are shown for each AC method, where both measures are calculated from hitting times of $100$ validation runs for \oea and AC methods. We also indicate the statistical significance in the empirical distributions of the hitting time ($^{****}$ for $p<0.001$, $^{***}$ for $p<0.01$, $^{**}$ for $<0.01$, and $^{*}$ for $p<0.05$) for each pair of the AC result and that of the \oea. The Mann–Whitney U test is applied with the Benjamini and Hochberg method for all 120 pairwise comparisons to control the false discovery rate. Runs are cut off at $50\,000$ evaluations if it does not hit the final target. The significant comparisons are colour-coded with respect to the relative improvement, where a darker colour signifies a more considerable improvement.}
}
\label{tab:result}

\begin{tabular}{|r||c|l|l|l|l||c|l|l|l|l|l|l|}
\hline
\multirow{2}{*}{\textbf{F}}	&	\multicolumn{5}{c||}{Relative improvement of ERT using ERT as the cost metric}	&\multicolumn{5}{c|}{Relative improvement of AUC using AUC as the cost metric}
\\
\cline{2-11}
	&\textbf{EA}	&	\textbf{GridSearch}	&	\textbf{Irace}	&	\textbf{MIP-EGO}	&	\textbf{MIES}		&\textbf{EA}	&	\textbf{GridSearch}	&	\textbf{Irace}	&	\textbf{MIP-EGO}&\textbf{MIES}\\	
\hline
1&665& -0.54$^{****}$ & -0.47$^{****}$ & -0.34$^{****}$ & \cellcolor[gray]{ 0.9473 } 0.01  &0.9987& -0.11$^{****}$ & -0.49$^{**}$ & -1.21$^{****}$ & -0.70  \\
2&5574& -0.64$^{****}$ & -0.12$^{****}$ &-0.03& \cellcolor[gray]{ 0.9279 } 0.05  &0.9514& -1.37$^{****}$ & -0.51$^{****}$ & -4.64$^{****}$ & \cellcolor[gray]{ 0.9480 } 0.25  \\
3&694& -0.62$^{****}$ & -0.53$^{****}$ & -0.78$^{****}$ & \cellcolor[gray]{ 0.9382 } 0.03  &0.9989& -0.11$^{****}$ & -1.12$^{****}$ &-0.57& -0.50  \\
4&344& -0.87$^{****}$ &-0.04& -0.30$^{****}$ & -0.11$^{*}$ &0.999& -0.06$^{****}$ &-0.66& -1.20$^{****}$ & -0.73$^{****}$ \\
5&598& -0.50$^{****}$ & -0.41$^{****}$ & -0.55$^{****}$ & \cellcolor[gray]{ 0.9338 } 0.04  &0.9987& -0.11$^{****}$ & -1.17$^{****}$ & -1.90$^{****}$ & -0.68$^{**}$ \\
6&271& -1.96$^{****}$ & -0.91$^{****}$ & -0.58$^{****}$ & -7.66$^{****}$ &0.9993& -0.11$^{****}$ &-0.43& -1.32$^{****}$ & -0.54$^{***}$ \\
7& Inf &---&---&---&Inf&0.8995&-0.33& \cellcolor[gray]{ 0.9364 } 1.69  & \cellcolor[gray]{ 0.9443 } 0.70  & \cellcolor[gray]{ 0.9481 } 0.24  \\
8&7926& \cellcolor[gray]{ 0.6006 } 0.78$^{****}$ & \cellcolor[gray]{ 0.6027 } 0.77$^{****}$ & -0.01$^{****}$ & \cellcolor[gray]{ 0.5651 } 0.86$^{****}$ &0.9938& \cellcolor[gray]{ 0.9479 } 0.25$^{****}$ & -0.37$^{****}$ & -0.65$^{****}$ & -0.75$^{****}$ \\
9&22670& \cellcolor[gray]{ 0.5675 } 0.85$^{****}$ & \cellcolor[gray]{ 0.5660 } 0.85$^{****}$ & \cellcolor[gray]{ 0.8831 } 0.15$^{****}$ & \cellcolor[gray]{ 0.6056 } 0.77$^{****}$ &0.9877& \cellcolor[gray]{ 0.9451 } 0.60$^{****}$ & \cellcolor[gray]{ 0.9466 } 0.42$^{****}$ & \cellcolor[gray]{ 0.9460 } 0.50$^{****}$ & \cellcolor[gray]{ 0.9467 } 0.41$^{****}$ \\
10& Inf &\cellcolor[gray]{0.5} Inf$^{****}$&\cellcolor[gray]{0.5} Inf$^{****}$&\cellcolor[gray]{0.5} Inf$^{****}$&\cellcolor[gray]{0.5} Inf$^{****}$&0.6362& \cellcolor[gray]{ 0.5070 } 54.86$^{****}$ & \cellcolor[gray]{ 0.5010 } 55.60$^{****}$ & \cellcolor[gray]{ 0.5018 } 55.50$^{****}$ & \cellcolor[gray]{ 0.5000 } 55.73$^{****}$ \\
11&2071& -0.29$^{****}$ & -0.35$^{****}$ & -0.45$^{****}$ & \cellcolor[gray]{ 0.8972 } 0.12$^{**}$ &0.9807& -0.36$^{****}$ &-1.05&-0.89& -0.93  \\
12&4691& -0.45$^{****}$ & -0.11$^{***}$ & -0.12$^{***}$ & \cellcolor[gray]{ 0.9244 } 0.06$^{*}$ &0.9601& -1.51$^{****}$ & -0.64$^{***}$ & -0.15$^{*}$ & \cellcolor[gray]{ 0.9489 } 0.13  \\
13&997& -0.83$^{****}$ & -0.14$^{*}$ & -0.70$^{****}$ &-0.05&0.9902& -0.57$^{****}$ &-0.8&-0.99& -1.26$^{****}$ \\
14&8171& \cellcolor[gray]{ 0.5094 } 0.98$^{****}$ & \cellcolor[gray]{ 0.5054 } 0.99$^{****}$ & \cellcolor[gray]{ 0.5089 } 0.98$^{****}$ & \cellcolor[gray]{ 0.5186 } 0.96$^{****}$ &0.596& \cellcolor[gray]{ 0.5649 } 47.69$^{****}$ & \cellcolor[gray]{ 0.5681 } 47.29$^{****}$ & \cellcolor[gray]{ 0.6532 } 36.75$^{****}$ & \cellcolor[gray]{ 0.6185 } 41.05$^{****}$ \\
15&6668& -1.46$^{****}$ & -1.17$^{****}$ & -0.50$^{****}$ & -1.23$^{****}$ &0.9404& -7.04$^{****}$ & -8.98$^{****}$ & -2.05$^{****}$ & -0.88$^{****}$ \\
16&9520& -2.05$^{****}$ & -0.97$^{****}$ & -0.43$^{****}$ & -0.23$^{****}$ &0.9174& -17.54$^{****}$ & \cellcolor[gray]{ 0.9467 } 0.41  & -12.97$^{****}$ & -1.78$^{****}$ \\
17&45964& -Inf$^{****}$ & -Inf$^{****}$ & -Inf$^{****}$ & -0.96$^{****}$ &0.6698& -48.43$^{****}$ & \cellcolor[gray]{ 0.9302 } 2.45$^{***}$ & -14.40$^{****}$ & -7.93$^{****}$ \\
18&130863& \cellcolor[gray]{ 0.5689 } 0.85$^{****}$ & \cellcolor[gray]{ 0.5886 } 0.80$^{****}$ & \cellcolor[gray]{ 0.6376 } 0.69$^{****}$ & \cellcolor[gray]{ 0.5660 } 0.85$^{****}$ &0.9432& \cellcolor[gray]{ 0.9350 } 1.86$^{***}$ & \cellcolor[gray]{ 0.9394 } 1.32  & \cellcolor[gray]{ 0.9495 } 0.06  & \cellcolor[gray]{ 0.9398 } 1.26$^{****}$ \\
19&9467& -525.05$^{****}$ & -Inf$^{****}$ & -1.72$^{****}$ & -0.72$^{****}$ &0.9899& -6.19$^{****}$ & \cellcolor[gray]{ 0.9500 } 0.00  & -4.96$^{****}$ & -0.37$^{**}$ \\
20&1460& -9.77$^{****}$ & -0.83$^{****}$ & -2.16$^{****}$ & -2.47$^{****}$ &0.9956& -1.26$^{****}$ & -0.20$^{****}$ & -0.25$^{****}$ & -0.68$^{****}$ \\
21&948& -12.75$^{****}$ & -3.37$^{****}$ & -3.66$^{****}$ & -3.95$^{****}$ &0.9965& -0.40$^{****}$ & -0.56$^{****}$ & -0.42$^{****}$ & -0.04$^{****}$ \\
22&3366& -2.60$^{****}$ & -0.11$^{****}$ & -0.57$^{****}$ &-0.15&0.9993& -0.02$^{****}$ & -0.05$^{****}$ &-0.06& -0.02$^{****}$ \\
23&3066& \cellcolor[gray]{ 0.9153 } 0.08$^{***}$ & -1.00$^{****}$ & -0.68$^{****}$ & -1.03$^{*}$ &0.9993& \cellcolor[gray]{ 0.9499 } 0.01$^{**}$ & -0.12$^{****}$ & -0.40$^{***}$ & -0.23$^{****}$ \\
24& Inf &\cellcolor[gray]{0.5} Inf$^{****}$&\cellcolor[gray]{0.5} Inf$^{****}$&\cellcolor[gray]{0.5} Inf$^{****}$&\cellcolor[gray]{0.5} Inf$^{****}$&0.9545& \cellcolor[gray]{ 0.9410 } 1.11$^{**}$ & \cellcolor[gray]{ 0.9385 } 1.42$^{****}$ & \cellcolor[gray]{ 0.9427 } 0.90  & \cellcolor[gray]{ 0.9362 } 1.71$^{****}$ \\
25&48946& \cellcolor[gray]{ 0.8512 } 0.22$^{*}$ & -Inf$^{****}$ & \cellcolor[gray]{ 0.6546 } 0.66$^{****}$ & \cellcolor[gray]{ 0.6999 } 0.56$^{****}$ &0.6953& \cellcolor[gray]{ 0.9494 } 0.07$^{**}$ & -0.00$^{****}$ & \cellcolor[gray]{ 0.9496 } 0.04  & -0.03$^{**}$ \\
\hline										
\multicolumn{2}{|c|}{\#improvements} &  8	&	6	&	6	&	13	&	-&		8	&	9	&	7	&	8
\\\hline
\end{tabular}
\end{table*}

The configuration space of the AC problem is $\Theta = \{\mu, \lambda, p_c, p_m\}$, where $\mu \in [100]$ is the parent population size, $\lambda \in [100]$ is the offspring population size, $p_c \in [0,1]$ is the crossover probability, and $p_m \in [0.005, 0.5]$ is the mutation rate. A positive crossover probability $p_c > 0$ requires $\mu > 1$; the configuration is considered infeasible otherwise. 
The results for the grid search are based on our work~\cite{Yeppsn2020}, where we used $\mu \in \{10,50,100\}$, $\lambda \in \{1,\mu/2, \mu\}$, $p_c\in \{0,0.5\}$, and $p_m=0.01$. Note that this is a considerably smaller search space, whose full enumeration requires only $18$ different configurations, which is much less than the budget allocated to the automated configuration techniques. We will nevertheless observe that for some problems none of the automated configurators could find hyperparameter settings that are equally good as those provided by this small grid search.

\subsection{Results Obtained by Automated Configuration}
\label{sec:resultOfAC}

The \oea with $p_m=1/n$ has shown competitive results in \cite{doerr2019benchmarking} for the PBO problems, so we use it as the baseline against which we compare the GAs obtained by the configurators. Note that this algorithm is part of the GA framework in Alg.~\ref{alg:GA} and could therefore be identified by the configuration methods.

Tab.~\ref{tab:config} lists the configurations of the \mlga obtained by the grid search and by the three AC methods.  
Tab.~\ref{tab:result} compares the performance of these configurations, by listing the ERT and AUC values of the \oea and the corresponding relative deviations of the configured GAs.   
More precisely, the ERT and AUC values of the AC methods result from using ERT and AUC as the cost metric, respectively. 
The relative improvement of ERT is computed as
(ERT$_{(1+1)\text{ EA}}$ - ERT) / ERT$_{(1+1)\text{ EA}}$, and the relative improvement of AUC is computed as (AUC - AUC$_{(1+1) \text{ EA}}$) / AUC$_{(1+1) \text{ EA}}$. 

\subsubsection{ERT Results}

For the \om-based problems F1 and F4-F6, configurations of the \mlga using crossover outperform the mutation-only GA with $\mu \ge 10$~\cite{Yeppsn2020}.  However, according to the values in Tab.~\ref{tab:result}, the \oea outperforms the configurations with $p_c>0$, relatively large $\mu$, and also relatively large $\lambda$. This observation matches our expectation because our previous study has shown that the \oea is efficient on \om.
Meanwhile, we observe an interesting configuration with $p_m < 0.01$, $\mu = \lambda =2$, and $p_c>0$ that achieves competitive ERT values against the \oea for F1.  This configuration ties well with the results on different $(\mu+1)$~GAs that were shown to outperform the \oea (and any mutation-based algorithm, in fact) in a series of recent works~\cite{PintoD18PPSN,CorusO18,sudholt2017crossover}. 

On F4, F6, F7, F13, F15-F17, and F19-22, none of the configurations returned by the AC methods was able to outperform the \oea, whereas 
on F9-10, F14, F18, F24, all AC methods find configurations that perform much better than the \oea. 

On \leadingones, a slightly better result compared to the ERT value $5\,574$ of the \oea is found by MIES.  This result is quite sensitive with respect to the mutation rate. When changing it from the MIES-suggestion of $p_m=0.005$ to $p_m=0.01$ we obtain an ERT value of $5\,829$.
MIES also obtains an improvement on F11, which corresponds to a $(2+2)$~GA with $p_m=0.0245$.
On F14, we already observed in~\cite{Yeppsn2020} that mutation-only GAs with $p_m = 1/n$ are inferior to other GA configurations with a larger offspring population size and higher mutation rate. As expected, all three methods easily suggest configurations that outperform the \oea by a great margin.

On F18 and F24, all configurators unanimously suggest fairly small values for the crossover probabilities.
For F25, however, GAs with $p_c>0.8$ show the (by far) best performance.

\subsubsection{AUC results}
Since we evaluate the AUC at each budget $[50\,000]$, the AUC values tend to be very close to $1$, especially for the GAs that require much fewer than $50\,000$ evaluations to find an optimal solution.

On 13 out of the 25 problems, none of the configuration methods is able to identify a hyperparameter setting that yields better AUC values than those of the \oea. In some, but clearly not all of these cases, the achieved AUC values are not much worse than those of the \oea. The largest improvements are obtained on functions F10, F14, F17, F18, F24, and F7 (in this order). In most of these cases all four configuration techniques found improvements over the \oea except for F17, where Irace is the only method finding an improvement and except for F7, where all three automated configuration techniques find an improvement but not the grid search (the inverse is true on F23 but the advantage of the grid search is fairly small, not statistically significant, and the algorithm a mutation-only (10+1)~GA with $p_m=1/n$, which is very close to the \oea). 

We next discuss the results for the functions for which large improvements over the \oea were obtained.

\begin{figure}[hb]
    \centering
     \includegraphics[width=0.95\linewidth]{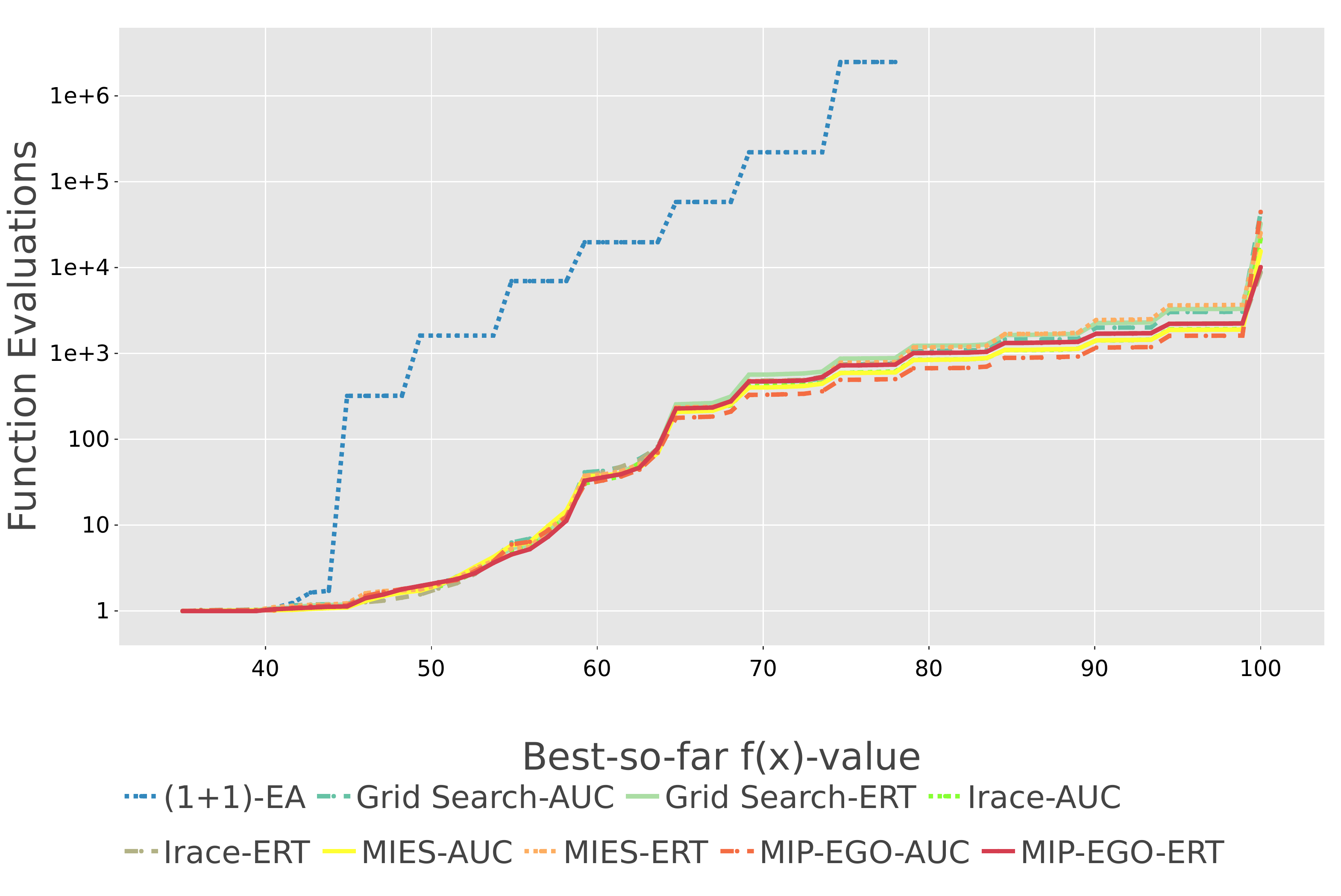}
    \caption{Fixed-target ERT values of the configurations suggested for F10. The suffix ``-ERT/AUC'' indicates which cost metric was used during the tuning.}
    \label{fig:F10}
\end{figure}

On \textbf{F10}, large crossover probabilities seem beneficial; all three automated configuration techniques return settings with $p_c>0.98$. The mutation rate seems to have less impact on the results, and the suggested settings vary from $8.5/n$ (Irace) to $41.3/n$ (MIES, which achieves the best AUC value). Interestingly, also two of the three configurations tuned for minimizing ERT have large crossover probabilities, with the exception of the one returned by MIES ($p_c=0.32$). The performances of all these configurations are very similar, as we can see in Fig.~\ref{fig:F10}. 

On \textbf{F14}, the best improvement is obtained by a mutation-only (100+50)~GA using $p_m=1/n$, closely followed by the Irace result, which is a (48+83)~GA using crossover probability $p_c=0.812$ and $p_m \approx 41/n$. We cannot observe any clear pattern in the results, and the four suggested configurations all differ quite a bit. 

For \textbf{F17}, as mentioned, only Irace finds a better configuration, which is also a (1+1)~EA, but using a slightly larger mutation rate of $p_m=1.9/n$ instead of $1/n$. 

For \textbf{F18}, all configurators return settings with small crossover probability $p_c<0.031$ and small mutation rate $p_m<0.01$, which indicates that local search methods may be more suitable for this problem than globally searching GAs. Interestingly, the performance of randomized local search is not very good (see~\cite{doerr2019benchmarking} for details), which suggests that a positive probability for escaping local optima via small jumps in the search space or via crossover are needed to be efficient on this problem.

For \textbf{F24}, no clear pattern can be observed in the suggested configurations, and also the crossover probabilities differ widely, from 0 for the grid search, values around 0.27 for MIP-EGO and MIES, to 0.7 for Irace.

On \textbf{F7}, the configuration obtained by Irace (which achieves the best improvement) is a $(1+\lambda)$~EA with $p_m \approx 3/n$, the one obtained by MIP-EGO is a $(2+20)$~GA with small crossover probability $p_c = 0.082$, and the one obtained by MIES is a (23+27)~GA with crossover probability $p_c = 0.672$. These results indicate that the fraction of configurations achieving better AUC value than the \oea may be fairly large. 

On the \leadingones problem F2, the $(2+1)$~GA with $p_m=0.008$ and $p_c=0.005$, found by MIES, yields a (small) improvement over the \oea, whereas the configurations found by the other methods perform worse. 

All in all, we find that on several problems the suggested configurations differ widely, far more than we would have expected and this across all four parameters. Analyzing the landscape of the AC problem suggests itself as an interesting follow-up study, e.g., using similar approaches to those suggested in~\cite{pushak2018algorithm,tanabe2020analyzing}.

\begin{figure*}[htb]
  \centering
  \includegraphics[width=0.8\linewidth]{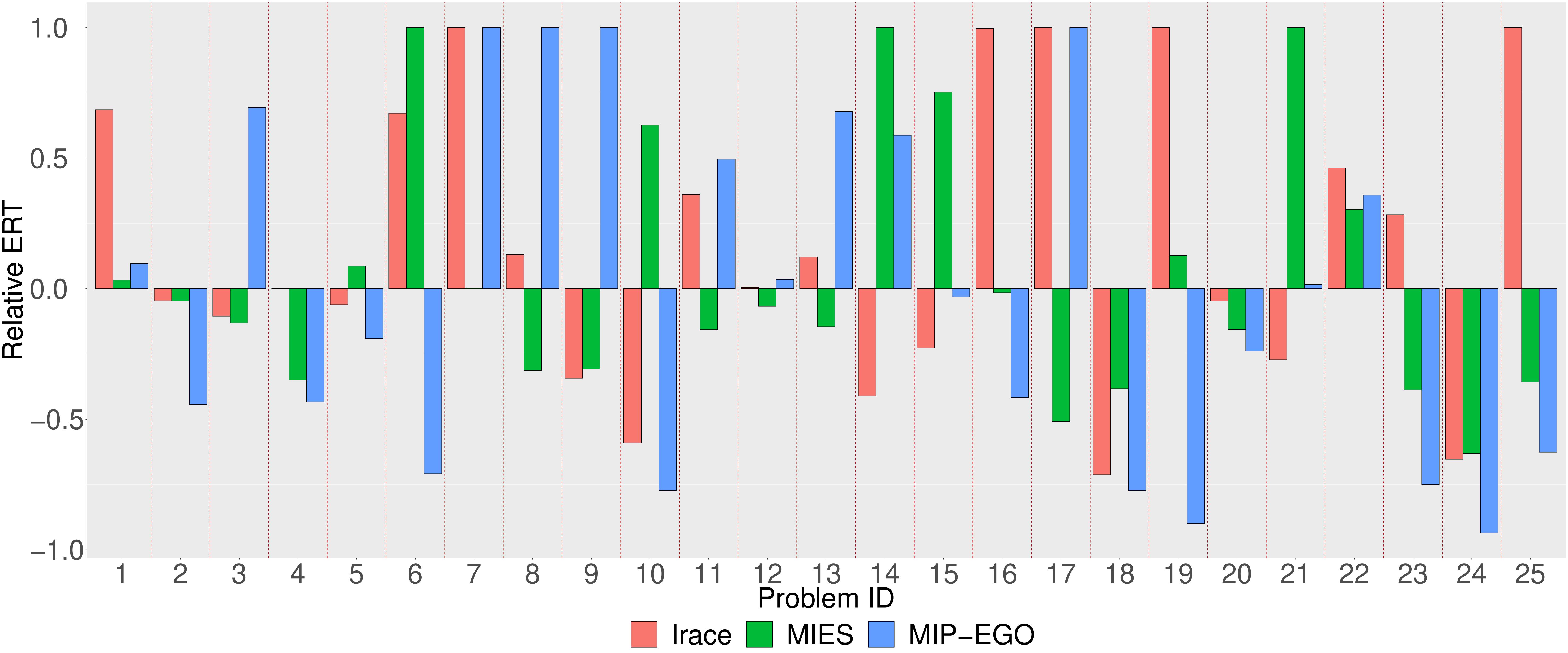}
  \caption{Relative ERT values of the GAs obtained by Irace, MIES, and MIP-EGO using ERT as the cost metric compared to the ERT of the GAs obtained by the same method when using AUC as the cost metric during the configuration process. Plotted values are (ERT$_{\text{using ERT}}$-ERT$_{\text{using AUC}}$)/ERT$_{\text{using AUC}}$, capped at $-1$ and $1$. Positive values therefore indicate that configurator obtains better results when configuring AUC.}
  \label{fig:AUC-ERT-PBO}
\end{figure*}

\begin{figure}[htb]
  \centering
  \includegraphics[width=\linewidth]{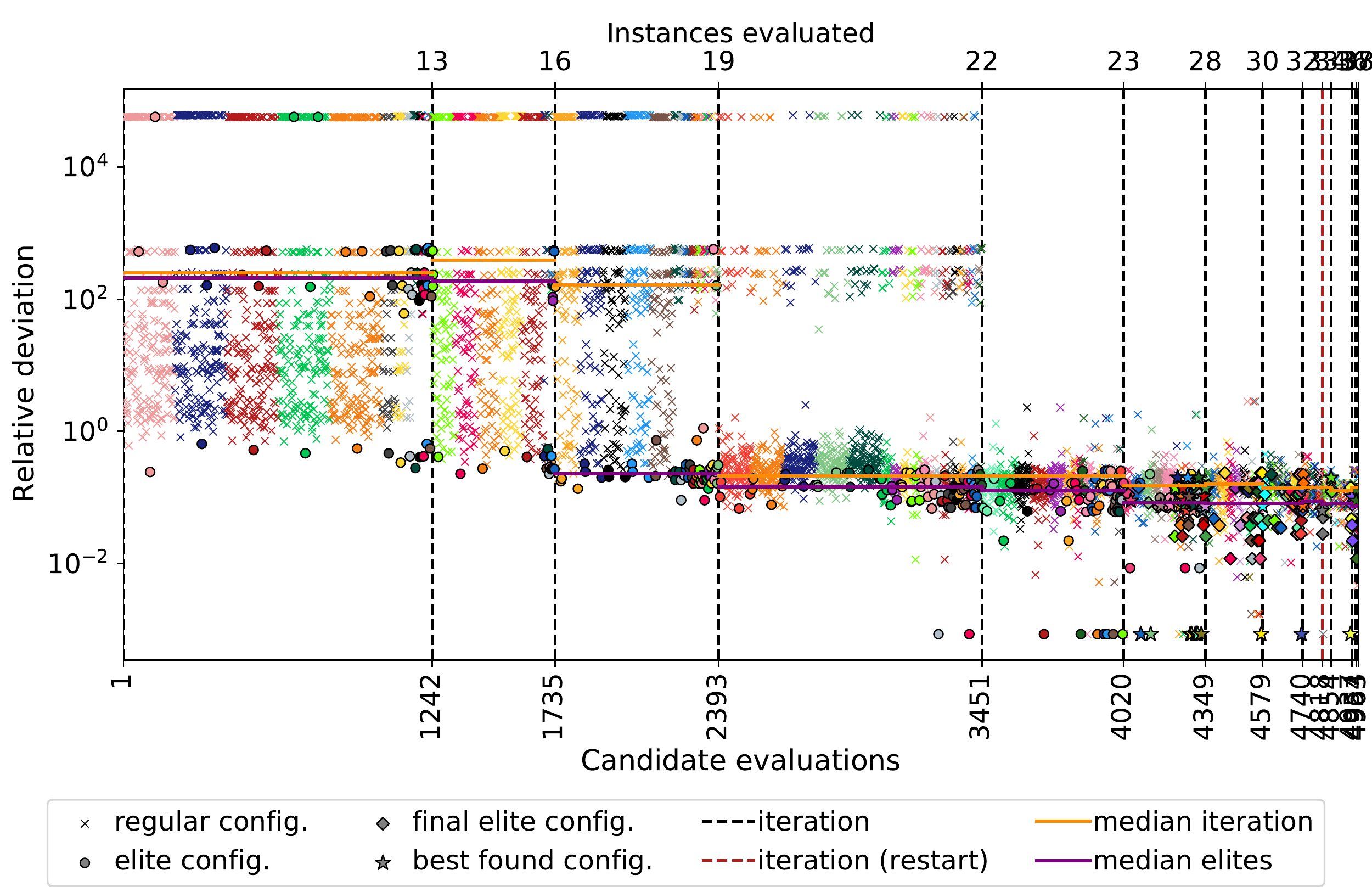}
  \caption{The relative deviation from the best-known ERT value of the GAs obtained during the configuration process of Irace for tuning the \mlga for \onemax in dimension $n=100$, with the objective to minimize the ERT for the optimum $f(x)=100$. The maximal number of configurations that can be tested by Irace is set to 5\,000. The figure is produced by the acviz tool, and illustrations for the details of plot representation can be found in Sec. 3 of~\cite{SouzaEtAl2020acviz}.}
  \label{fig:Irace-OM-ERT}
\end{figure}

\subsection{Discussions on the Configurators' Performance}
\label{sec:tuning}
We now compare the performance of the three automated AC methods. The last row of Tab.~\ref{tab:result} summarizes for how many settings each method was able to find configurations that outperform the \oea. These numbers are rather balanced between the different methods, with the notable exception of the minimizing ERT objective, for which MIES suggested 13 improvements, compared to 6-8 improvements found by the other methods. MIES also suggested the best configurations in most of the cases, but, for several of these, the improvements over the \oea are rather minor. This shows that barely counting such cases does not do justice to the complex behavior observed in Tab.~\ref{tab:result}, from which we cannot derive a clear winning configurator. We can nevertheless make a few observations.
 
\subsubsection{Handling conditional parameter spaces} 
We easily see from Tab.~\ref{tab:config} that Irace is the only method that obtains mutation-only GAs, and in all of these cases it returns a $(1+\lambda)$~EA. We recall that setting $\mu=1$ requires to set $p_c=0$; the configuration is infeasible otherwise.
This advantage of Irace lies in its handling of conditional parameters: Irace samples non-conditional parameters first, and samples conditional parameters only if the condition is satisfied. MIP-EGO and MIES, in contrast, sample parameter values from independent distributions and give penalties to infeasible settings. 
With this strategy, the two methods can avoid infeasible candidates, but the probability of sampling feasible conditional candidates may be too small.
For example, MIP-EGO can find a configuration with $\mu=2$ and $p_c = 0.0065$ on F16 in Tab.~\ref{tab:config}, but it cannot obtain the competitive configuration of  $(1+\lambda)$~mutation-only GA because the probability of sampling $\mu=1$ and $p_c = 0$ simultaneously is too small.

\subsubsection{Impact of the cost metric} We have already observed that MIES obtains better configurations for more problems when using ERT as the cost metric. For AUC, in contrast, Irace finds more configurations that improve over the \oea, which can be explained as follows. In the first few iterations, AUC is able to differentiate the performance of two poor configurations if both fail to find the final target, whereas the ERT value will be infinite and thereby incomparable in this case. Hence, using AUC as the cost metric, Irace could learn to avoid evaluating those poor configurations in the following iterations. It is worth noting that such an observation is also supported by a case study of Irace~\cite{caceres2017experimental}, in which the authors discovered that Irace would spend too much time on poor configurations if the mean running time is taken as the cost metric. As a solution, the adaptive capping strategy~\cite{ParamILS} is introduced to Irace in this work. Interestingly, this discussion connotes that the AUC metric realizes a similar effect as with adaptive capping for minimizing the running time of an optimization algorithm.
This behavior also indicates that the choice of the cost metric might be a factor to consider when choosing which AC technique to apply.

For an algorithm that cannot hit the target in all runs,
the variance of its ERT values can be high due to the uncertain success rate. 
Besides, ERT cannot distinguish algorithms that cannot hit the target in any runs, even though their performance may differ in terms of results for other targets.
This shortcoming is mitigated when tuning for large AUC, since this performance metric also takes into account the hitting times for easier targets.

Fig.~\ref{fig:Irace-OM-ERT} plots the relative deviations from the best-known ERT value of the configurations obtained during one run of Irace when using ERT as the cost metric. 
We observe that many configurations show large relative deviation values, which stem from GAs that cannot hit the optimum within the given budget. These configurations do not provide much useful information, since they all look equally bad for the configurator. 

\subsection{The Choice of the Cost Metric}
\label{sec:ChoiceOfMetric}

\begin{figure}[tb]
\begin{framed}
    \centering
    \includegraphics[width=\linewidth]{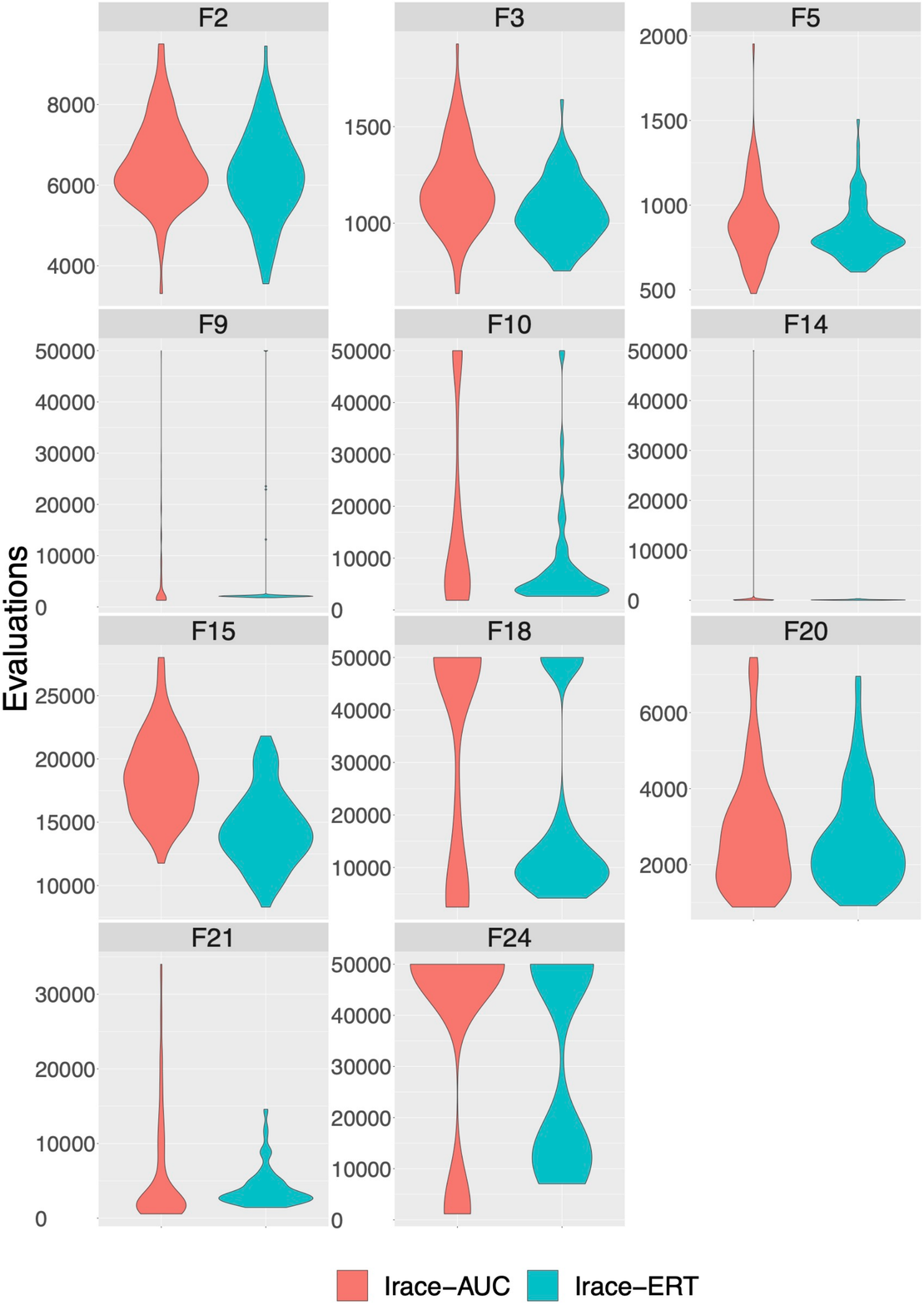}
    \end{framed}
    \caption{Violin plots of first hitting times for the configurations found by Irace when tuning for ERT and AUC, respectively. Only showing results for problems on which \textbf{Irace-ERT outperforms Irace-AUC}. 
    Results are from the 100 independent validation runs. Targets are listed in Tab.~\ref{tab:targets-f}, and the configurations of the GAs can be found in Tab.~\ref{tab:config}. For each run, values are capped at the budget $50\,000$ if the algorithm cannot find the target.}
    \label{fig:vilion-AUC-ERT-worse}
\end{figure}

\begin{figure}[tb]
    \centering
    \begin{framed}
    \includegraphics[width=\linewidth]{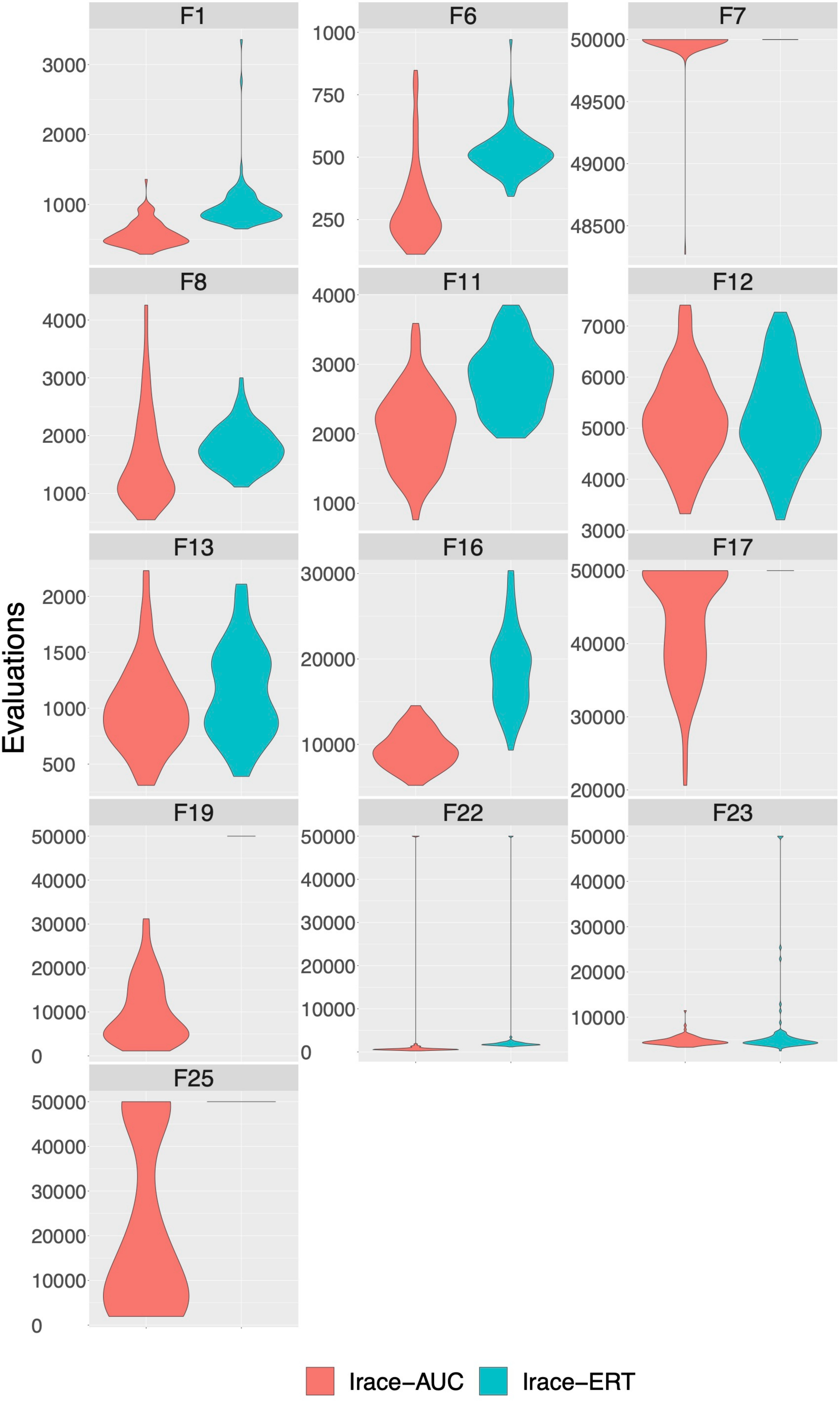}
    \end{framed}
    \caption{Violin plots of first hitting times for the configurations found by Irace when tuning for ERT and AUC, respectively, for problems on which \textbf{Irace-AUC outperforms Irace-ERT}. 
    Results are from the 100 independent validation runs. Targets are listed in Tab.~\ref{tab:targets-f}, and the configurations of the GAs can be found in Tab.~\ref{tab:config}. For each run, values are capped at the budget $50\,000$ if the algorithm cannot find the target.}
    \label{fig:vilion-AUC-ERT-better}
\end{figure}

\begin{figure}[tb]
  \centering
  \includegraphics[width=0.75\linewidth]{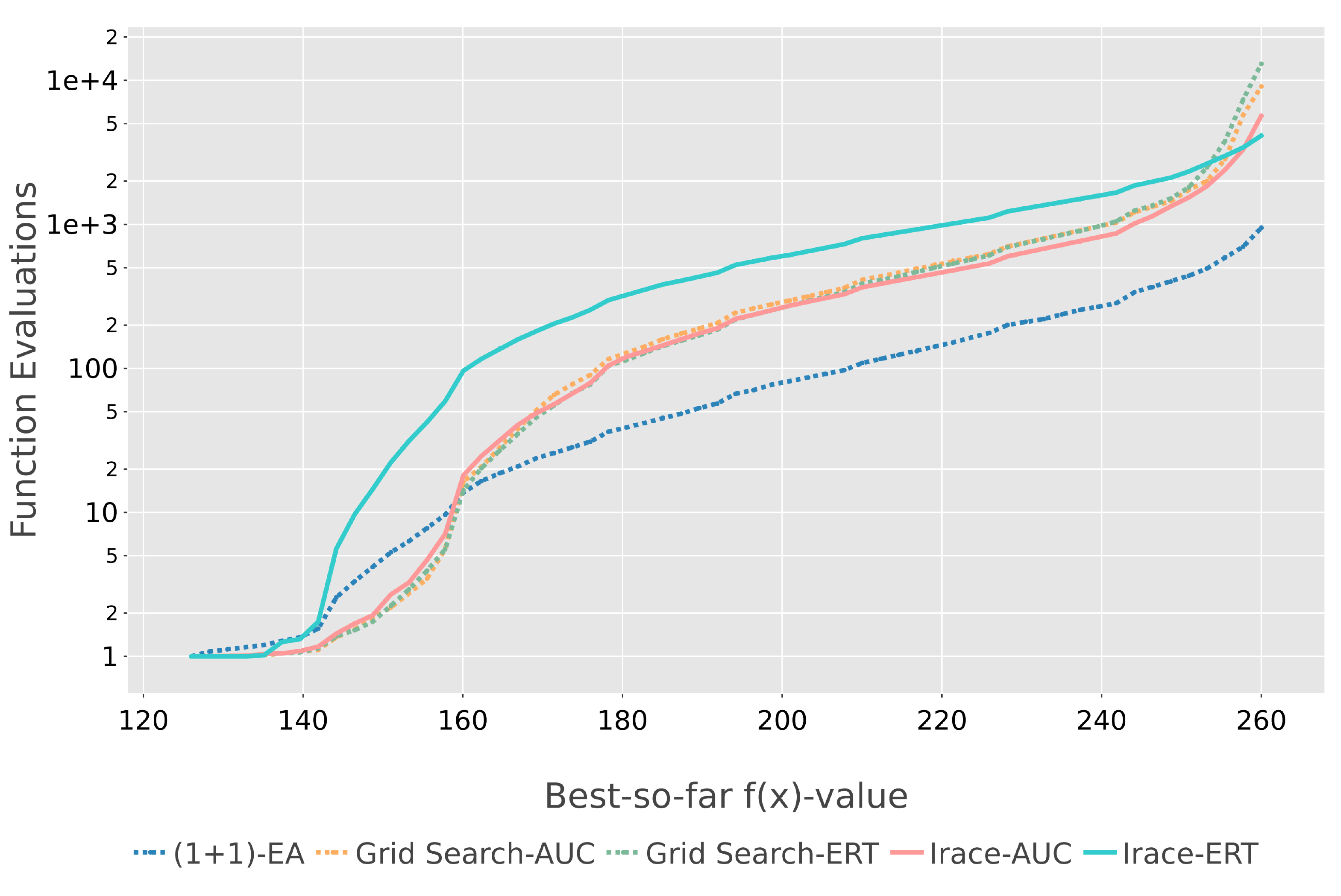}
  \caption{Fixed-target ERT values of the GAs listed in Tab.~\ref{tab:config} for F21.}
  \label{fig:ERT-F21}
\end{figure}

\begin{figure}[htb]
  \centering
  \includegraphics[width=0.75\linewidth]{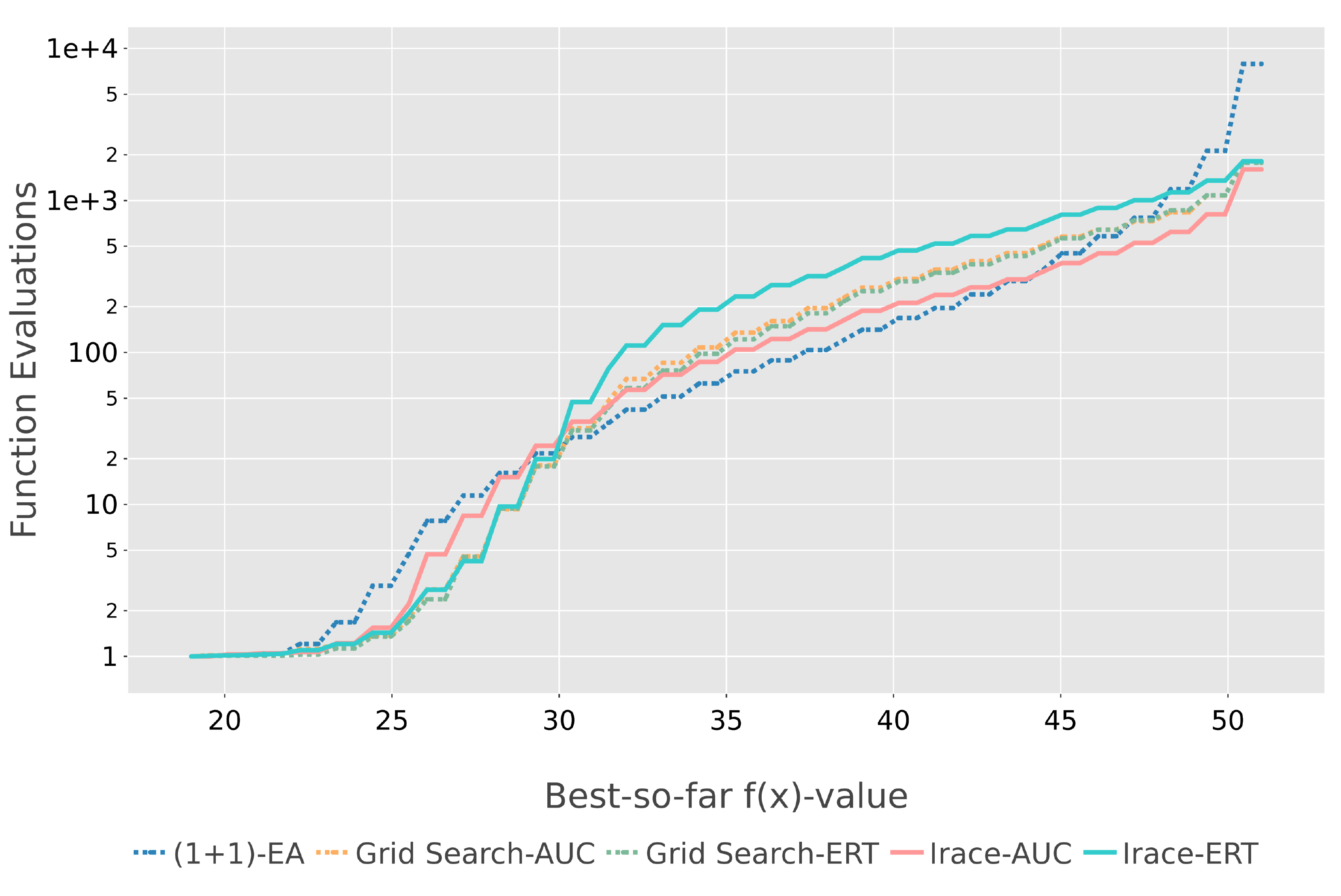}
  \caption{Fixed-target ERT values of the GAs listed in Tab.~\ref{tab:config} for F8.}
  \label{fig:ERT-F8}
\end{figure}

\begin{figure}[t]
  \centering
  \includegraphics[width=0.75\linewidth]{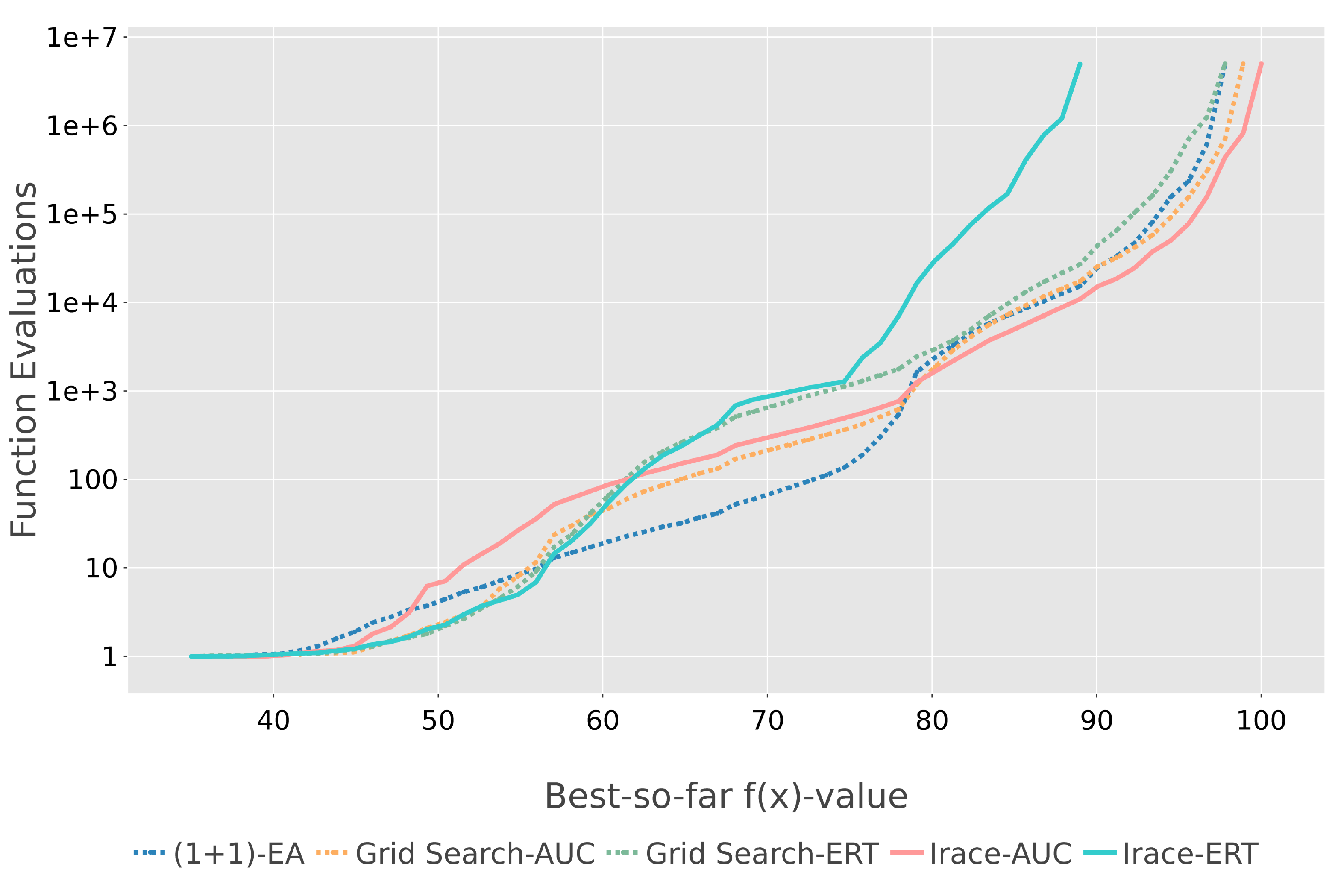}
  \caption{Fixed-target ERT values of the GAs listed in Tab.~\ref{tab:config} for F7.}
  \label{fig:ERT-F7}
\end{figure}

We now evaluate how well configurations that are obtained by tuning for the AUC cost metric perform in terms of ERT. Fig.~\ref{fig:AUC-ERT-PBO} summarizes these result, by plotting the relative advantage of the  configurations tuned for AUC, compared to those that were explicitly tuned for ERT. More precisely, we plot (ERT$_{\text{using ERT}}$-ERT$_{\text{using AUC}}$)/ERT$_{\text{using AUC}}$, so that positive values indicate that tuning for AUC gave better ERT values than the configurations obtained when tuning for ERT. We see that this is the case for 13, 12, and 9 out of the 25 problems when using Irace, MIP-EGO, and MIES, respectively.  

We now zoom into the results obtained by Irace. We abbreviate by ``Irace-ERT'' (``Irace-AUC'') the configurations obtained when using ERT (AUC) as the cost metric. 
For the problems on which the ERT of Irace-AUC was worse than that of Irace-ERT, we plot in Fig.~\ref{fig:vilion-AUC-ERT-worse} violin plots for the running times of the 100 validation runs. We observe that F15 is the only problem where Irace-ERT significantly outperforms Irace-AUC. 
On F2-3, F5, and F20, we observe that the result of most runs of Irace-AUC and Irace-ERT are close, but the variances of the results of Irace-AUC are higher than for Irace-ERT. 
On the remaining problems, we observe high variances for the result of both Irace-ERT and Irace-AUC. 
Irace-ERT finds the configurations with fewer \textit{unsuccessful} runs, which makes sense because the number of \textit{unsuccessful} runs significantly affects the ERT value. 
However, AUC does not only consider the evaluations needed to hit the final target, so we observe more \textit{unsuccessful} runs and \textit{competitive} partial runs for Irace-AUC, i.e., in cases of F21 and F24.

Although Irace-AUC does not obtain better ERT values than Irace-ERT, it can still provide valuable insights concerning the resulting configurations and performance profiles.
Fig.~\ref{fig:ERT-F21} plots the fixed-target ERT values of the GAs obtained by Irace-ERT and Irace-AUC for F21. 
We observe that the result of Irace-ERT outperforms the result of Irace-AUC for the final target $f(x)=260$. 
However, for the long period when $f(x) < 258$, Irace-AUC performs better. 

We plot in Fig.~\ref{fig:vilion-AUC-ERT-better} the violin plots of the running times for the problems where Irace-AUC obtains better ERT values than Irace-ERT. The advantage of Irace-AUC is significant on several problems, i.e., F1, F6, F11, F16, and F22. 
Moreover, Irace-ERT cannot find the final targets of F7, F17, F19, and F25 within the cutoff time, whereas Irace-AUC hits the targets in some (F7, F17, and F25) or all (F19) of the runs.

Fig.~\ref{fig:ERT-F8} plots the fixed-target result of different GAs on F8. 
Compared to Irace-ERT, we observe that Irace-AUC outperforms the other algorithm at the final target and also exhibits advantages over other algorithms in most of the optimization process.
Fig.~\ref{fig:ERT-F7} plots the fixed-target result of different GAs on F7. 
The figure shows that Irace-AUC is the only one that hits the optimum $f(x) = 100$, and the best-found fitness of Irace-ERT is less than $90$. 
We observe that none of the GAs shown in the figure hits the optimum in all runs (because the ERT values are larger than the cutoff time of $50\,000$ function evaluations). 

The results of Irace-AUC and Irace-ERT on the PBO problems reveal the questions of how the cost metric affects the performance of Irace for different configuration tasks for future study. We study in the following the impact of the cutoff time concerning the behavior of Irace on \onemax and \leadingones.

\begin{figure}[tb]
  \centering
  \includegraphics[width=\linewidth]{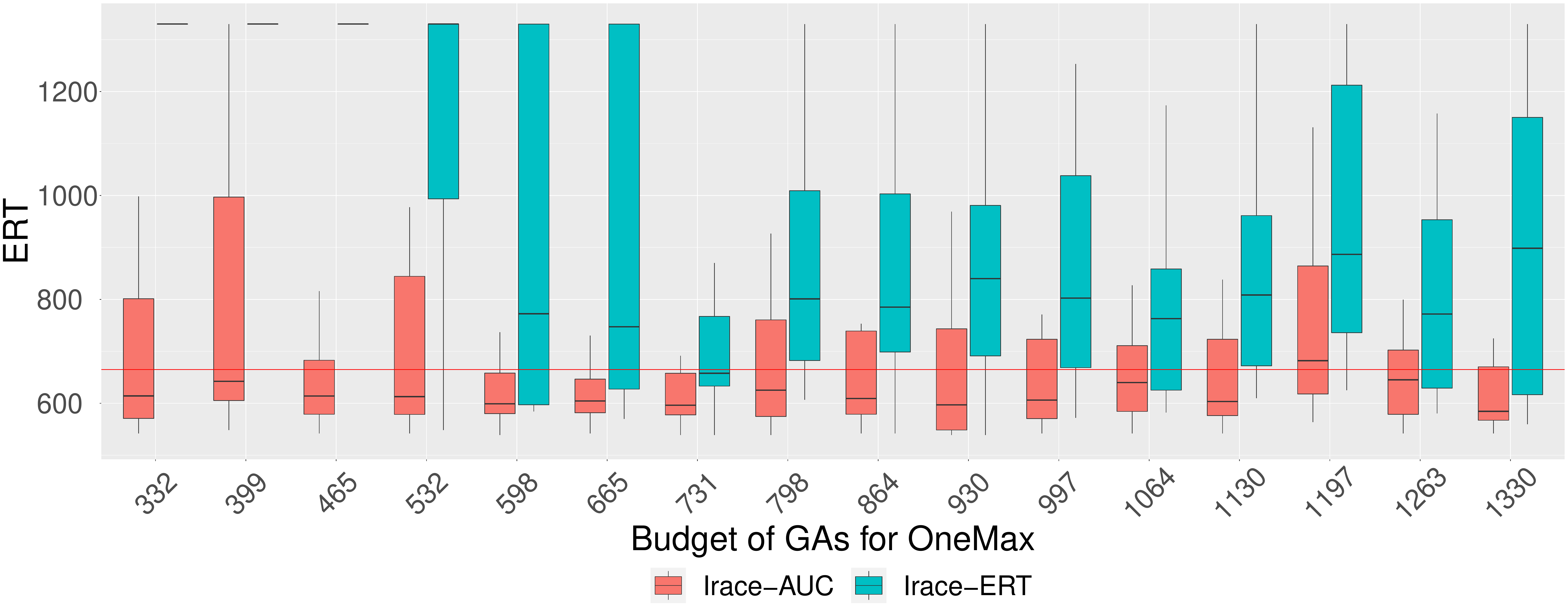}
  \includegraphics[width=\linewidth]{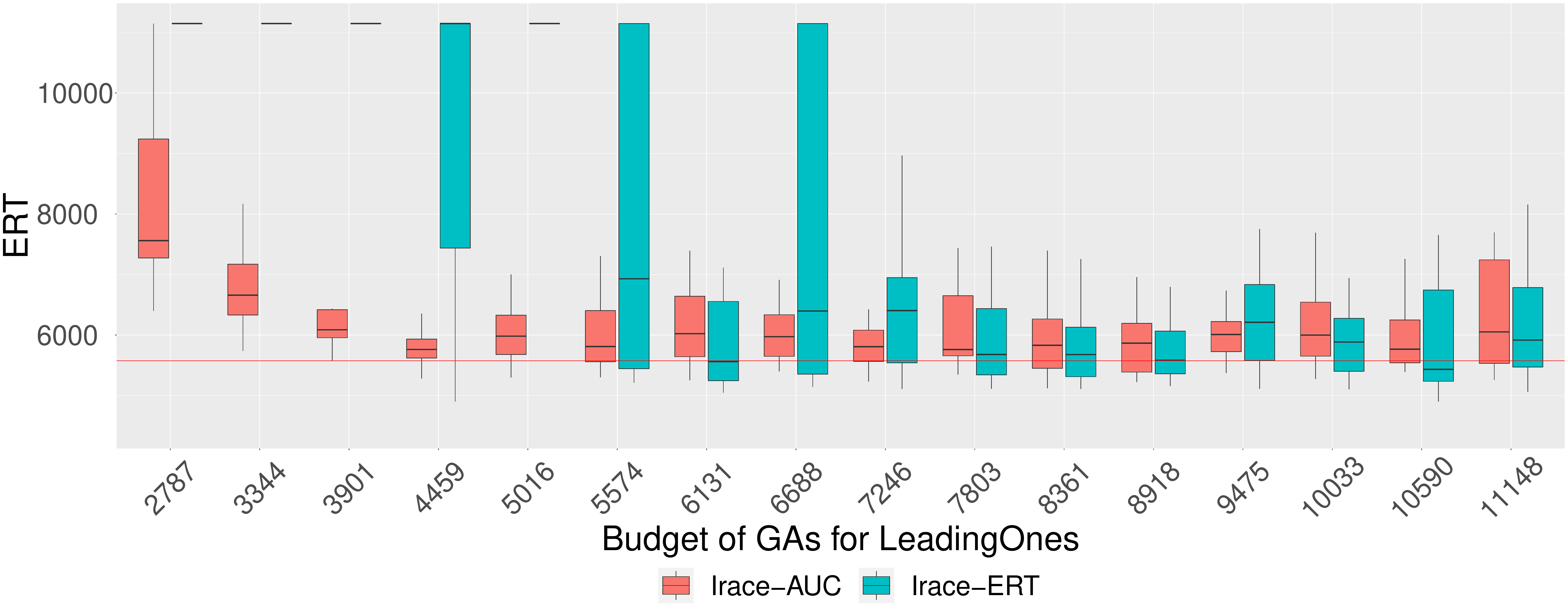}
  \caption{ERT values ($y$-axis) of the GAs obtained by Irace for \onemax and \leadingones in dimension $n=100$, for different cutoff time $B$ that the GAs can spend to find the optimum ($x$-axis). Showing results for $B \in \{(0.5 + 0.1t) \text{ERT}_{(1+1) \text{ EA}} \mid  t \in [0..15]\}$. For comparison, the ERT values of the \oea are plotted by horizontal red lines. Results are for the best found configurations obtained from $r=20$ independent runs of Irace, and each of the ERT values is with respect to 100 independent validation runs. }
  \label{fig:diffB}
\end{figure}

\subsubsection{Sensitivity with respect to the cutoff time} 
Inspired by the result in Fig.~\ref{fig:ERT-F7}, we study the sensitivity of ERT and AUC with respect to the cutoff time of the GAs. To this end, we consider the set $\{(0.5 + 0.1t) \times \text{ERT}_{\text{\oea}} \mid t \in [0..15] \}$ of 16 different cutoff times. 
For each of these cutoff time, for each of the two cost metrics (AUC and ERT), and for each of \textbf{F1} and \textbf{F2}, we run Irace 20 independent times with the same configuration budget of 5\,000 target runs (where each target run corresponds again to ten independent runs of the respective \mlga configuration). 
Fig.~\ref{fig:diffB} plots ERT values of the GAs obtained this way (as before, each ERT value is based on 100 independent validation runs). For comparison, the red line indicates the performance of the \oea. 

On \onemax, we observe that Irace-ERT cannot find promising configurations when the cutoff time of the GAs is too small to hit the optimum. This is the case for cutoff times smaller than $665$. 
However, Irace-AUC can work with small budgets that are not sufficient to hit the optimum. 
Even with cutoff times larger than $665$, Irace-AUC still obtains better ERT values than Irace-ERT.
Similarly, the result for \leadingones shows that Irace-ERT cannot find promising configurations with insufficient cutoff times. 
Still, Irace-AUC performs well across all cutoff times. 

Overall, we thus see that tuning with respect to AUC is much less sensitive with respect to the cutoff time.

\begin{figure}[tb]
  \centering
   \subfigure[F1]{
  \includegraphics[width=0.4\linewidth]{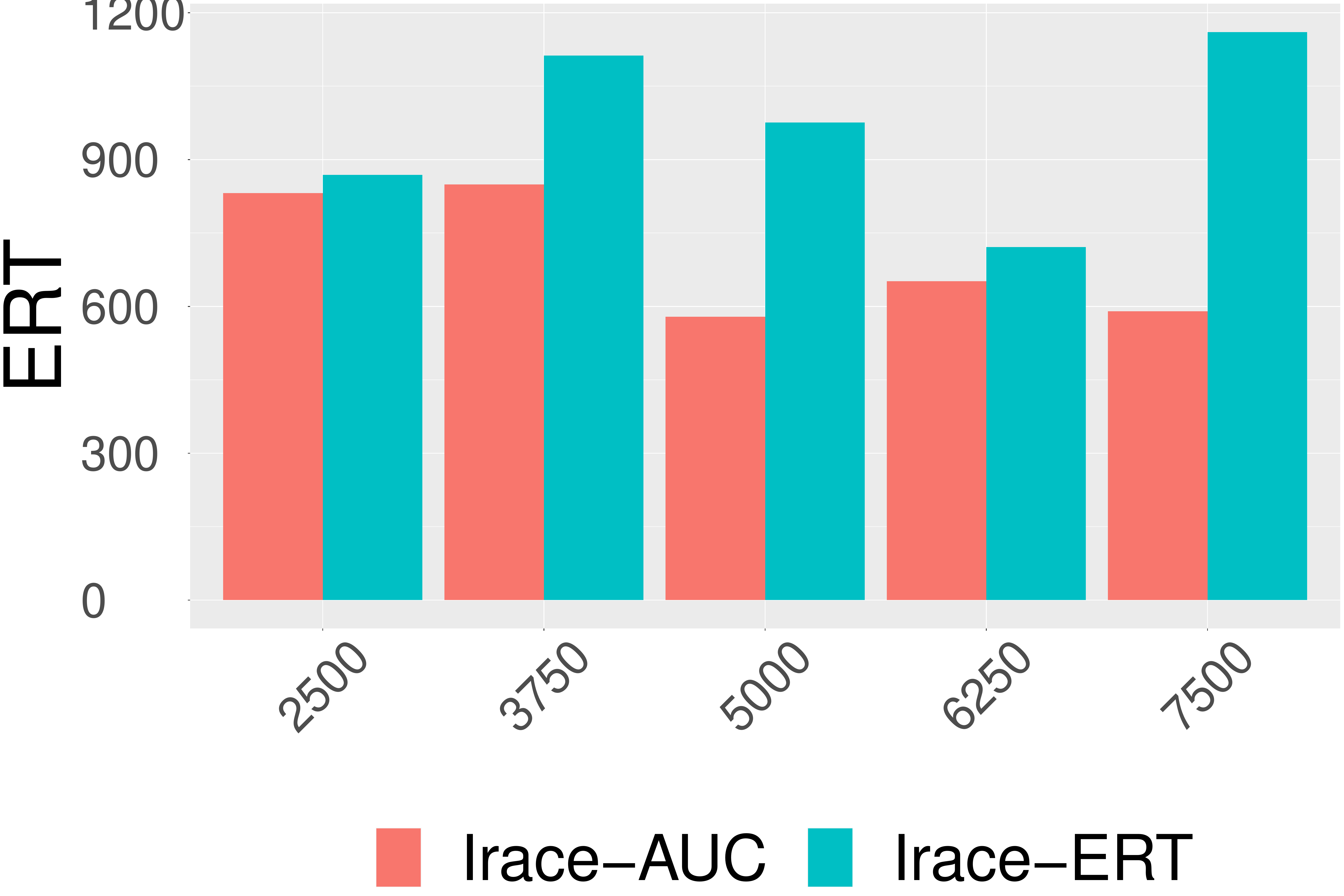}
  }
  \subfigure[F2]{
  \includegraphics[width=0.4\linewidth]{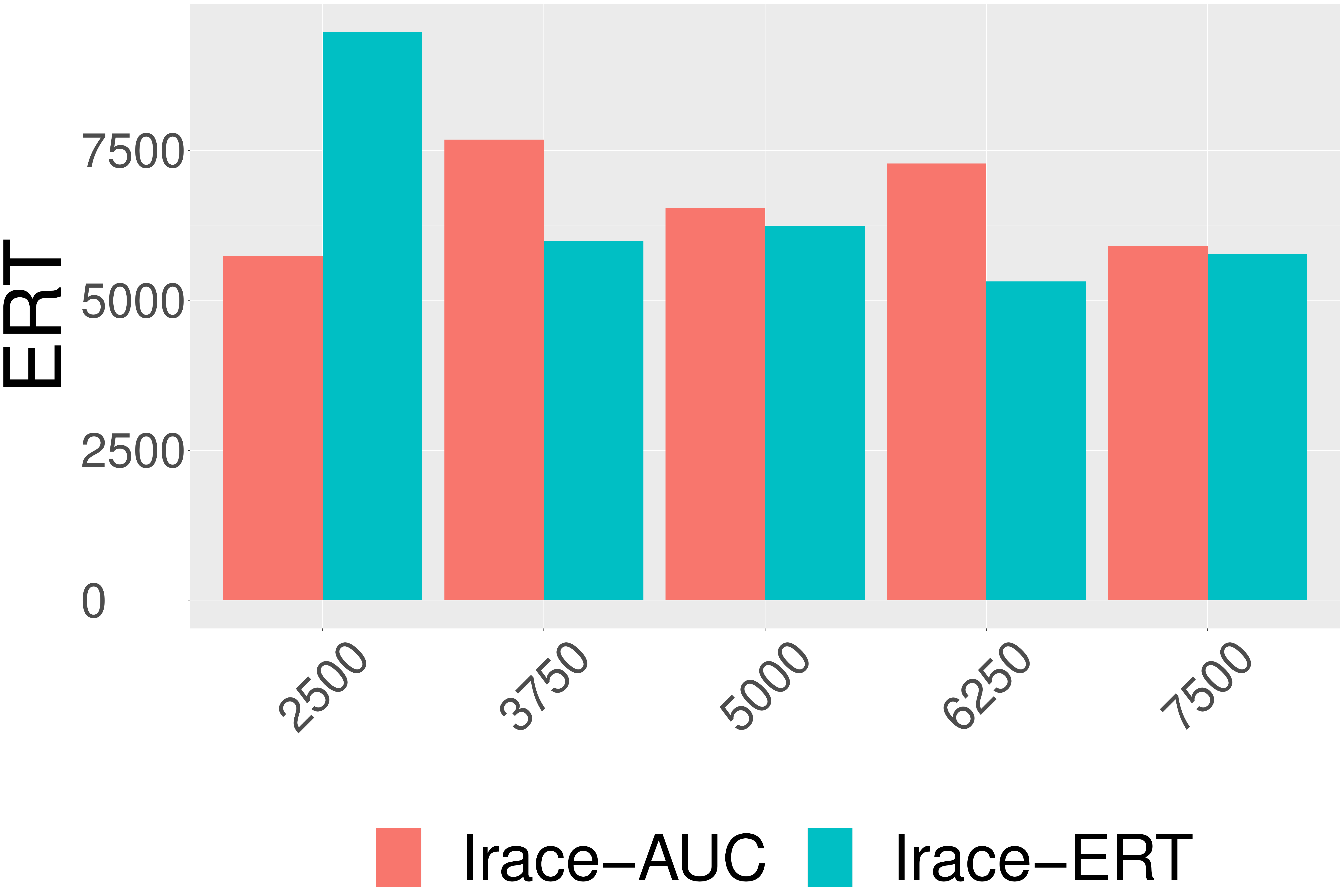}
  }
  \subfigure[F8]{
  \includegraphics[width=0.4\linewidth]{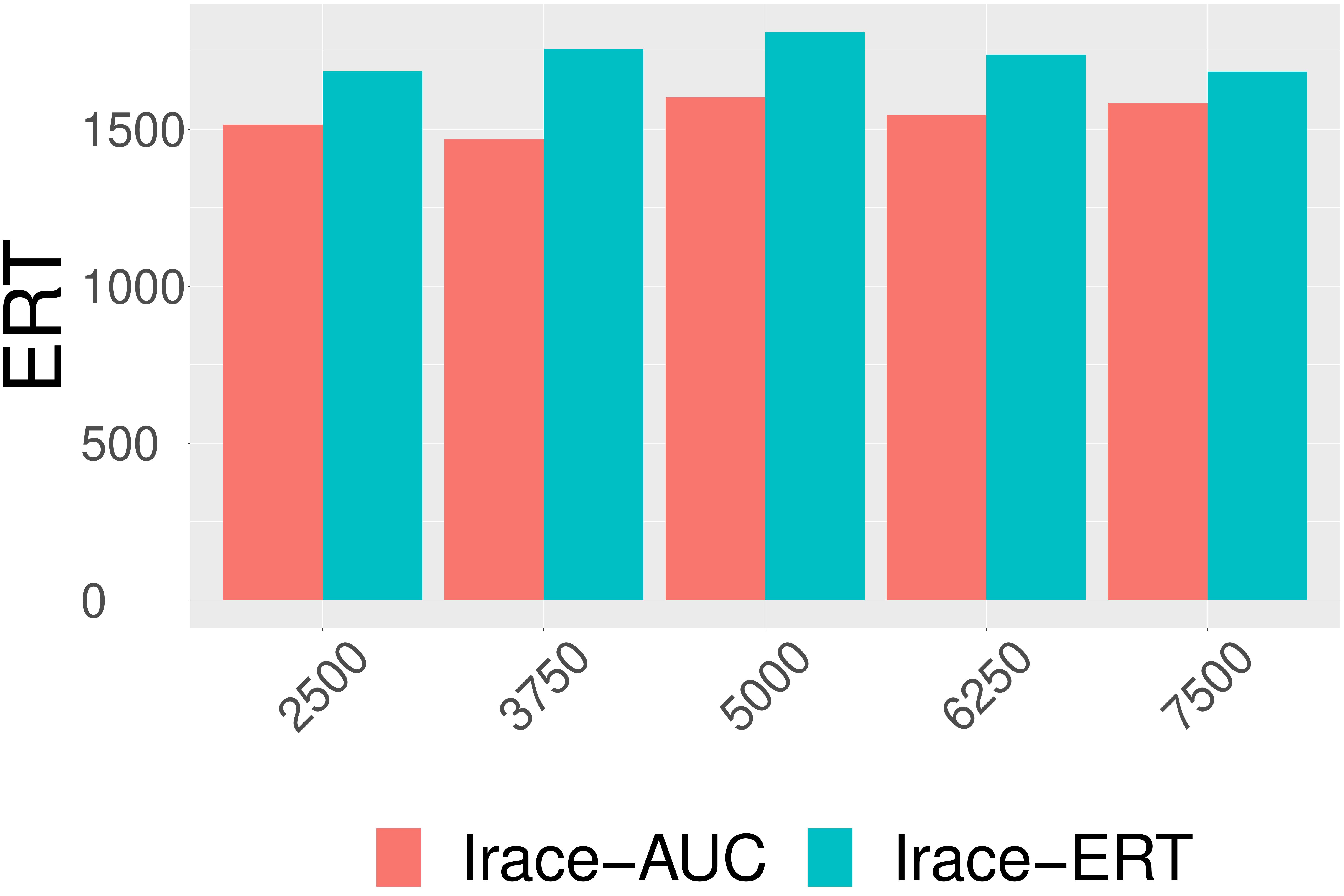}
  }
  \subfigure[F9]{
  \includegraphics[width=0.4\linewidth]{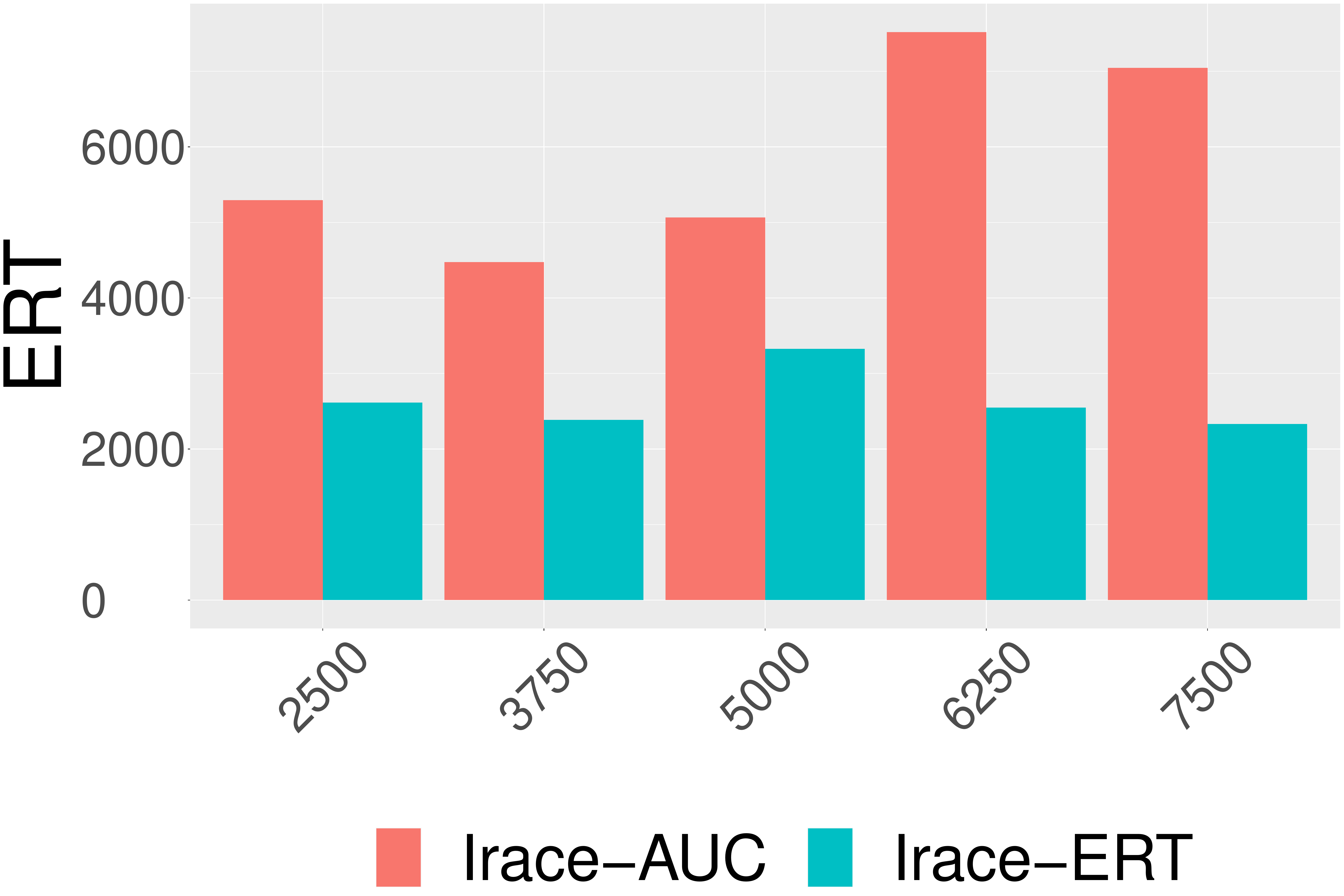}
  }
 \subfigure[F10]{
  \includegraphics[width=0.4\linewidth]{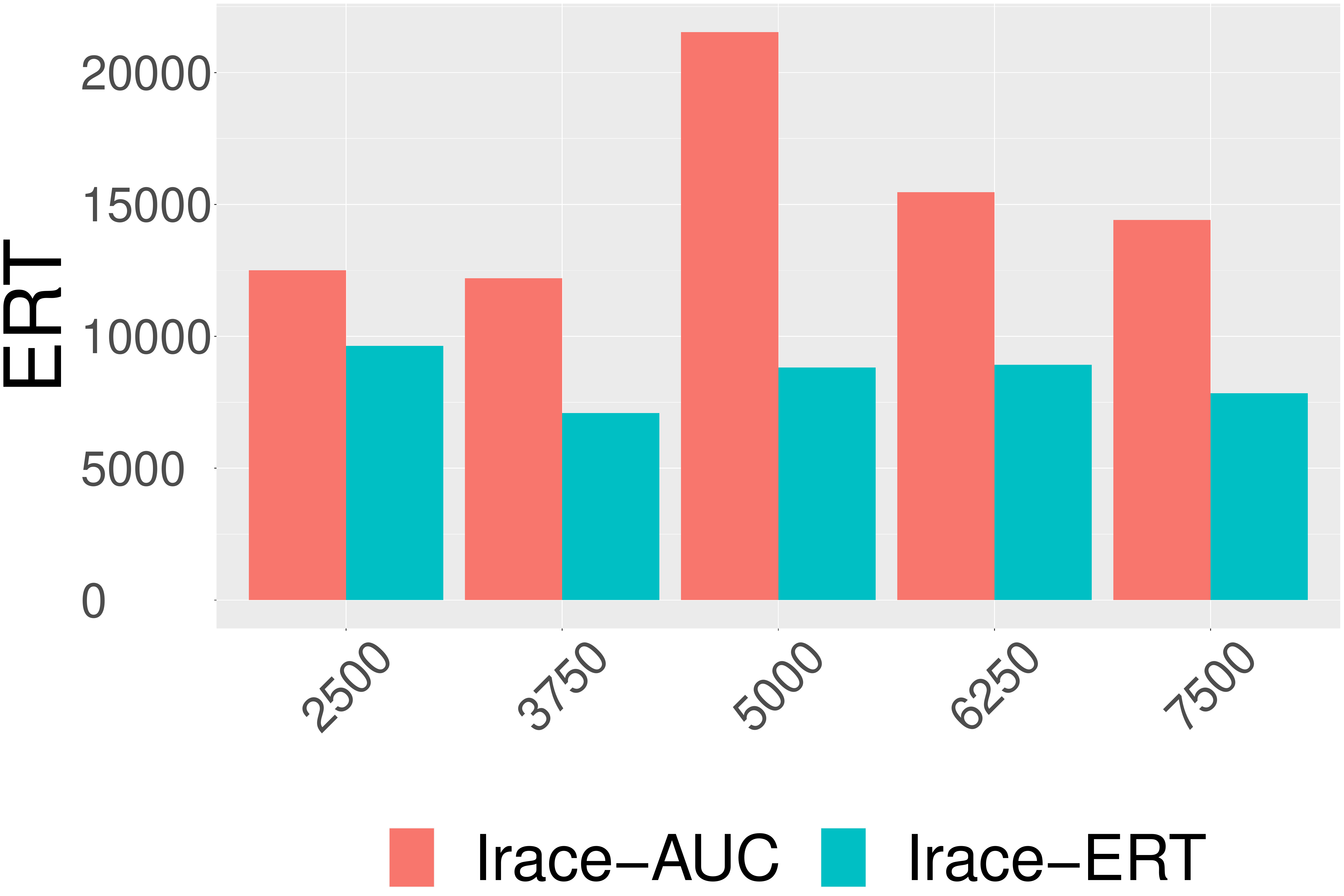}
  }
  \subfigure[F19]{
  \includegraphics[width=0.4\linewidth]{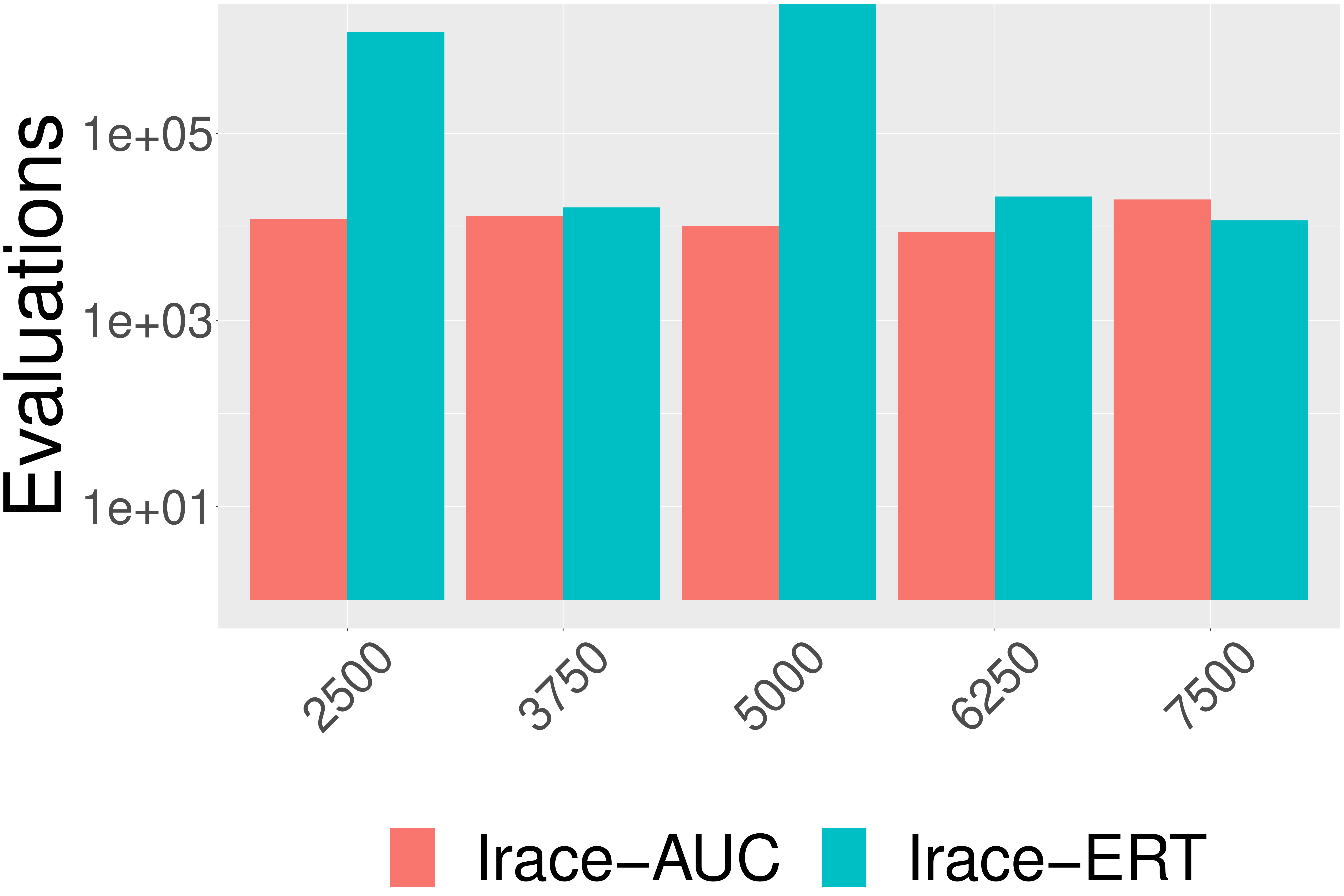}
  }
  \subfigure[F20]{
  \includegraphics[width=0.4\linewidth]{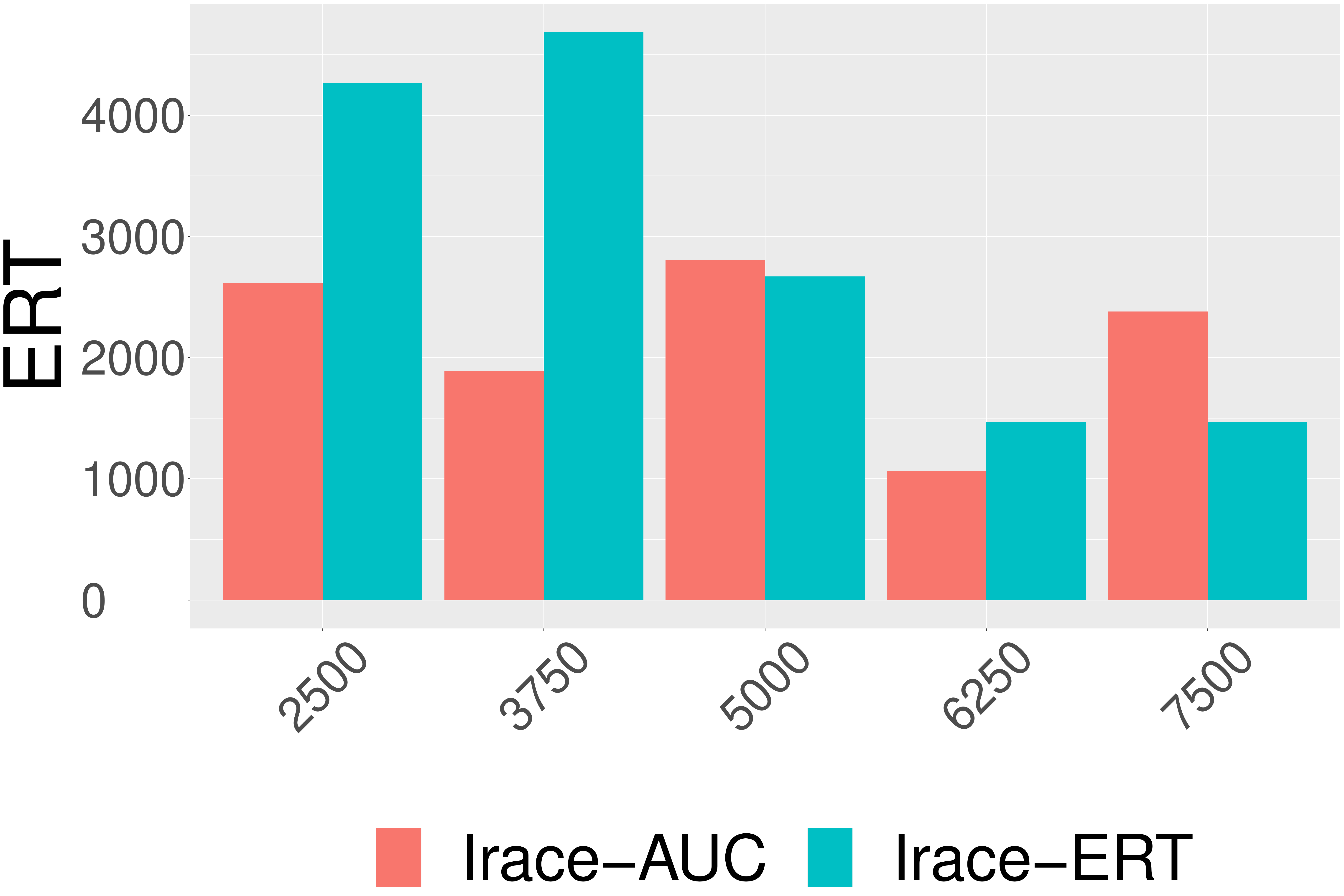}
  }\subfigure[F21]{
  \includegraphics[width=0.4\linewidth]{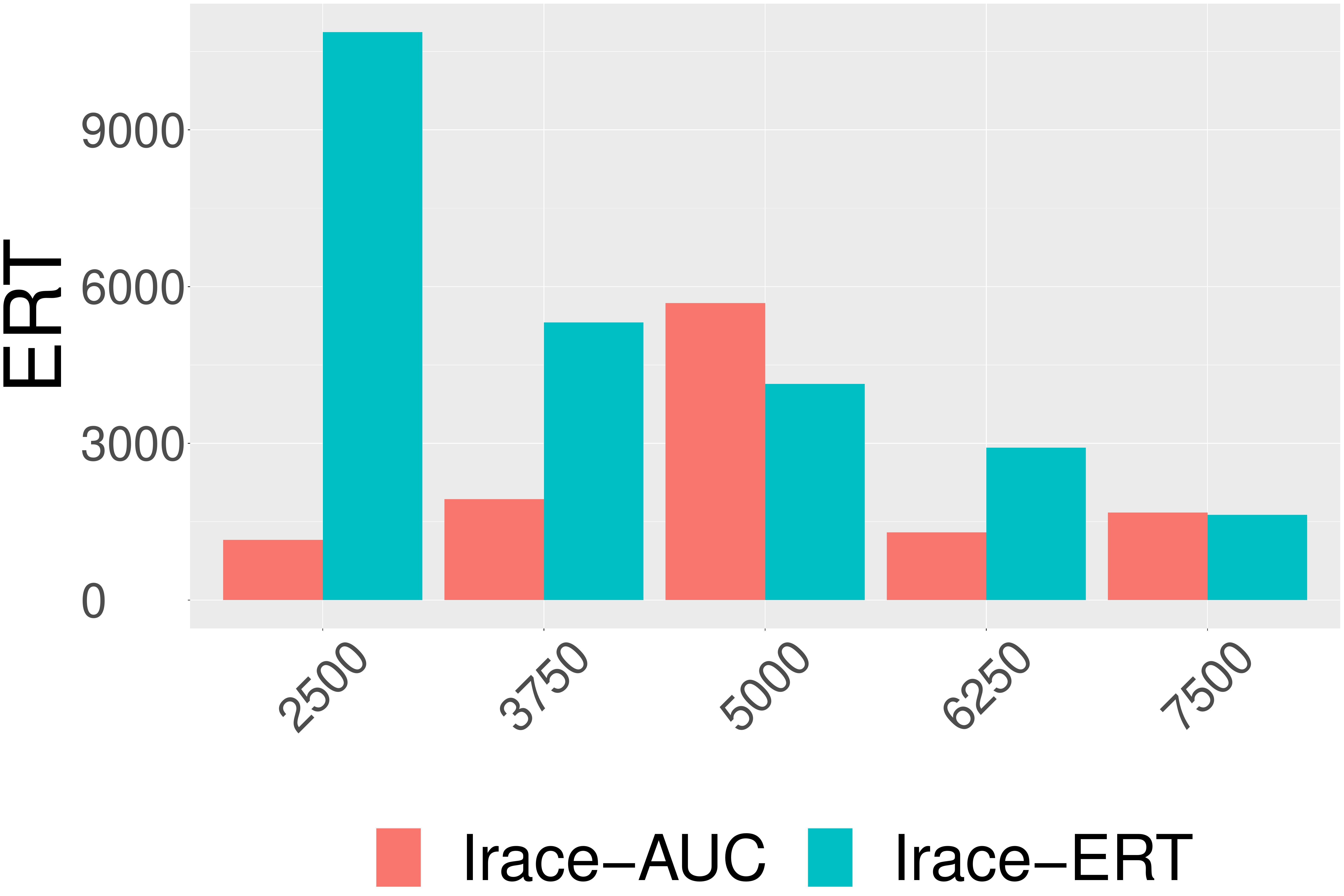}
  }
  \caption{ERT values ($y$-axis) of the GAs obtained by Irace with different configuration budgets $B_T$ (the number of configurations that Irace can test, $x$-axis). Results are for $B_T \in \{(0.5 + 0.25t) 5\,000 \mid t \in [0..4]\}$. Each ERT value is for the 100 validation runs of the configuration suggested by Irace after a single run, i.e., one for each budget.}
\label{fig:diff-Irace-B}
\end{figure}

\subsubsection{Sensitivity with respect to the configuration budget of Irace} 
We also analyze the sensitivity of the results with respect to the configuration budget, i.e., the number of target runs that the configurator can perform before it suggests a configuration. We use Irace for this purpose. Fig.~\ref{fig:diff-Irace-B} plots the ERT values of the configurations suggested by Irace, for 8 selected problems from the PBO suite. Interestingly, the ERT values are not monotonically decreasing, as one might have expected, at least for the configurations that are explicitly tuned for small ERT. Tuning for AUC gave the best ERT values for F1, 8, 19, 20, and 21.

\section{Discussion and Conclusions}
\label{sec:conclude}
In this paper, we have analyzed the performance of a family of $(\mu+\lambda)$~GAs in which offspring are generated by applying either crossover (with probability $p_c$) or mutation (probability $1-p_c$).  
Four different configuration methods have been applied for finding promising configurations of GAs: grid search and three automated techniques. 

The experimental results show that mutation-only GAs usually benefit from small parent population size. 
On the contrary, crossover-based GAs require sufficient population sizes. 
On our PBO problem set, the \oea outperforms the other tested GAs on \onemax, \leadingones, and some of their W-model extensions. 
However, crossover can be beneficial for the W-model extensions with epistasis and ruggedness, concatenated trap, and NK-landscapes.

We have also investigated the performance of the AC methods, Irace, MIP-EGO, and MIES. Irace is the only method that could find $(1+\lambda)$~mutation-only GAs, highlighting the importance of handling conditional parameter spaces appropriately.

We also observed that the cost metric used as tuning objective has a major impact on the performance of the AC methods. When using ERT, the AC methods cannot obtain useful information from configurations that cannot hit the optimum. But not only for these cases we observed that tuning for AUC gave better ERT values than when directly tuning for ERT. 

Our results raise several interesting topics for future research. Apart from analyzing more closely the configurations obtained for each problem, and the sensitivity of the performance with respect to the different parameters, we also plan to study in what sense our observation that tuning for AUC can help finding better configurations for ERT generalizes to other algorithm families and/or problems. A bi-objective (or even multi-objective, if considering different performance measures) optimization process might be able to balance advantages of the different cost metrics. Such an approach would tie well with recent works in the context of algorithm selection~\cite{bossek2020multi,JesusLDP20,JesusPL21}.

Our results have also demonstrated that none of the AC methods clearly outperforms all others, suggesting to either combine them or to develop guidelines that can help users select the most suitable configuration technique for their concrete problem at hand. Finally, we have also observed that in several cases none of the techniques could find configurations that outperform or perform on par with the \oea, which may indicate improvement potential for these configuration methods. 

\section*{Acknowledgments}

We thank Manuel L{\'{o}}pez{-}Ib{\'{a}}{\~{n}}ez for pointing us to the acviz tool~\cite{SouzaEtAl2020acviz}.
We thank Thomas Weise for pointing out a mistake in the previous definition of the Ising model problems F19-21. 

Our work was supported by the Chinese scholarship council (CSC No. 201706310143), 
by the Paris Ile-de-France Region, by ANR-11-LABX-0056-LMH (LabEx LMH), 
and by COST Action CA15140.
\bibliographystyle{abbrv}
\bibliography{references}

\begin{thebibliography}{10}

\bibitem{AmineGECCO}
A.~Aziz{-}Alaoui, C.~Doerr, and J.~Dr{\'{e}}o.
\newblock Towards large scale automated algorithm design by integrating modular
  benchmarking frameworks.
\newblock In {\em Proc. of Genetic and Evolutionary Computation Conference
  (GECCO'21)}, pages 1365--1374. ACM, 2021.

\bibitem{back1994parallel}
T.~B{\"a}ck.
\newblock Parallel optimization of evolutionary algorithms.
\newblock In {\em In Proc. of Parallel Problem Solving from Nature (PPSN'94)},
  pages 418--427. Springer, 1994.

\bibitem{baeck96}
T.~B\"{a}ck.
\newblock {\em Evolutionary algorithms in theory and practice: Evolution
  strategies, evolutionary programming, genetic algorithms}.
\newblock Oxford University Press, Inc., USA, 1996.

\bibitem{back1991survey}
T.~B{\"a}ck, F.~Hoffmeister, and H.-P. Schwefel.
\newblock A survey of evolution strategies.
\newblock In {\em Proc. of International Conference on Genetic Algorithms
  (ICGA'91)}. Citeseer, 1991.

\bibitem{bartz2005sequential}
T.~Bartz-Beielstein, C.~W. Lasarczyk, and M.~Preu{\ss}.
\newblock Sequential parameter optimization.
\newblock In {\em Proc. of Congress on Evolutionary Computation (CEC'05)},
  pages 773--780. IEEE, 2005.

\bibitem{birattari2010f}
M.~Birattari, Z.~Yuan, P.~Balaprakash, and T.~St{\"u}tzle.
\newblock F-race and iterated f-race: An overview.
\newblock In {\em Experimental Methods for the Analysis of Optimization
  Algorithms}, pages 311--336. Springer, 2010.

\bibitem{bossek2020anytime}
J.~Bossek, P.~Kerschke, and H.~Trautmann.
\newblock Anytime behavior of inexact {TSP} solvers and perspectives for
  automated algorithm selection.
\newblock In {\em Proc. of Congress on Evolutionary Computation (CEC'20)},
  pages 1--8. IEEE, 2020.

\bibitem{bossek2020multi}
J.~Bossek, P.~Kerschke, and H.~Trautmann.
\newblock A multi-objective perspective on performance assessment and automated
  selection of single-objective optimization algorithms.
\newblock {\em Applied Soft Computing}, 88:105901, 2020.

\bibitem{PintoD18PPSN}
E.~{Carvalho Pinto} and C.~Doerr.
\newblock A simple proof for the usefulness of crossover in black-box
  optimization.
\newblock In {\em Proc. of Parallel Problem Solving from Nature (PPSN'18)},
  pages 29--41. Springer, 2018.

\bibitem{CorusO18}
D.~Corus and P.~S. Oliveto.
\newblock Standard steady state genetic algorithms can hillclimb faster than
  mutation-only evolutionary algorithms.
\newblock {\em {IEEE} Transactions on Evolutionary Computation},
  22(5):720--732, 2018.

\bibitem{DangD19}
N.~Dang and C.~Doerr.
\newblock Hyper-parameter tuning for the {(1} + (\emph{{\(\lambda\)},
  {\(\lambda\)}})) {GA}.
\newblock In {\em Proc. of Genetic and Evolutionary Computation Conference
  (GECCO'19)}, pages 889--897. ACM, 2019.

\bibitem{DeJong75}
K.~A. De~Jong.
\newblock {\em An analysis of the behavior of a class of genetic adaptive
  systems}.
\newblock PhD thesis, University of Michigan, Ann Arbor, MI, USA, 1975.

\bibitem{SouzaEtAl2020acviz}
M.~de~Souza, M.~Ritt, M.~L{\'o}pez-Ib{\'a}{\~n}ez, and L.~P. C{\'a}ceres.
\newblock Acviz: A tool for the visual analysis of the configuration of
  algorithms with irace.
\newblock {\em Operations Research Perspectives}, 8:100186, 2021.

\bibitem{doerr2017fast}
B.~Doerr, H.~P. Le, R.~Makhmara, and T.~D. Nguyen.
\newblock Fast genetic algorithms.
\newblock In {\em Proc. of Genetic and Evolutionary Computation Conference
  (GECCO'17)}, pages 777--784. ACM, 2017.

\bibitem{IOHprofiler}
C.~Doerr, H.~Wang, F.~Ye, S.~van Rijn, and T.~B{\"a}ck.
\newblock {IOHprofiler}: A benchmarking and profiling tool for iterative
  optimization heuristics.
\newblock {\em arXiv e-prints:1810.05281}, Oct. 2018.

\bibitem{doerr2019benchmarking}
C.~Doerr, F.~Ye, N.~Horesh, H.~Wang, O.~M. Shir, and T.~B{\"a}ck.
\newblock Benchmarking discrete optimization heuristics with {IOHprofiler}.
\newblock {\em Applied Soft Computing}, 88:106027, 2020.

\bibitem{eggensperger2019pitfalls}
K.~Eggensperger, M.~Lindauer, and F.~Hutter.
\newblock Pitfalls and best practices in algorithm configuration.
\newblock {\em Journal of Artificial Intelligence Research}, 64:861--893, 2019.

\bibitem{ELSAYED20111877}
S.~M. Elsayed, R.~A. Sarker, and D.~L. Essam.
\newblock Multi-operator based evolutionary algorithms for solving constrained
  optimization problems.
\newblock {\em Computers \& Operations Research}, 38(12):1877 -- 1896, 2011.

\bibitem{abs-1807-02811}
P.~I. Frazier.
\newblock A tutorial on {B}ayesian optimization.
\newblock {\em arXiv preprint arXiv:1807.02811}, 2018.

\bibitem{Goldberg89}
D.~E. Goldberg.
\newblock {\em Genetic algorithms in search, optimization and machine
  learning}.
\newblock Addison-Wesley Longman Publishing Co., Inc., USA, 1st edition, 1989.

\bibitem{grefenstette1986optimization}
J.~J. Grefenstette.
\newblock Optimization of control parameters for genetic algorithms.
\newblock {\em IEEE Transactions on Systems, Man, and Cybernetics},
  16(1):122--128, 1986.

\bibitem{hall2020analysis}
G.~T. Hall, P.~S. Oliveto, and D.~Sudholt.
\newblock Analysis of the performance of algorithm configurators for search
  heuristics with global mutation operators.
\newblock In {\em Proc. of the Genetic and Evolutionary Computation Conference
  (GECCO'20)}, pages 823--831, 2020.

\bibitem{HallOS22impact}
G.~T. Hall, P.~S. Oliveto, and D.~Sudholt.
\newblock On the impact of the performance metric on efficient algorithm
  configuration.
\newblock {\em Artif. Intell.}, 303:103629, 2022.

\bibitem{SMAC}
F.~Hutter, H.~H. Hoos, and K.~Leyton-Brown.
\newblock Sequential model-based optimization for general algorithm
  configuration.
\newblock In {\em Proc. of Learning and Intelligent Optimization (LION'11)},
  pages 507--523. Springer, 2011.

\bibitem{ParamILS}
F.~Hutter, H.~H. Hoos, K.~Leyton-Brown, and T.~St\"{u}tzle.
\newblock Param{ILS}: An automatic algorithm configuration framework.
\newblock {\em Journal of Artificial Intelligence Research}, 36:267--306, 2009.

\bibitem{Yoon22}
{Hyun-Sook Yoon} and {Byung-Ro Moon}.
\newblock An empirical study on the synergy of multiple crossover operators.
\newblock {\em IEEE Transactions on Evolutionary Computation}, 6(2):212--223,
  April 2002.

\bibitem{JesusLDP20}
A.~D. Jesus, A.~Liefooghe, B.~Derbel, and L.~Paquete.
\newblock Algorithm selection of anytime algorithms.
\newblock In {\em Proc. of Genetic and Evolutionary Computation Conference
  (GECCO'20)}, pages 850--858. ACM, 2020.

\bibitem{JesusPL21}
A.~D. Jesus, L.~Paquete, and A.~Liefooghe.
\newblock A model of anytime algorithm performance for bi-objective
  optimization.
\newblock {\em J. Glob. Optim.}, 79(2):329--350, 2021.

\bibitem{LehreW12}
P.~K. Lehre and C.~Witt.
\newblock Black-box search by unbiased variation.
\newblock {\em Algorithmica}, 64:623--642, 2012.

\bibitem{li2013mixed}
R.~Li, M.~T. Emmerich, J.~Eggermont, T.~B{\"a}ck, M.~Sch{\"u}tz, J.~Dijkstra,
  and J.~H. Reiber.
\newblock Mixed integer evolution strategies for parameter optimization.
\newblock {\em Evolutionary Computation}, 21(1):29--64, 2013.

\bibitem{lopez2016Irace}
M.~L{\'o}pez-Ib{\'a}{\~n}ez, J.~Dubois-Lacoste, L.~P. C{\'a}ceres,
  M.~Birattari, and T.~St{\"u}tzle.
\newblock The irace package: Iterated racing for automatic algorithm
  configuration.
\newblock {\em Operations Research Perspectives}, 3:43--58, 2016.

\bibitem{lopez2014automatically}
M.~L{\'o}pez-Ib{\'a}nez and T.~St{\"u}tzle.
\newblock Automatically improving the anytime behaviour of optimisation
  algorithms.
\newblock {\em European Journal of Operational Research}, 235(3):569--582,
  2014.

\bibitem{MitchellHF93}
M.~Mitchell, J.~H. Holland, and S.~Forrest.
\newblock When will a genetic algorithm outperform hill climbing?
\newblock In {\em Proc. of Neural Information Processing Systems Conference
  (NIPS'93)}, pages 51--58, 1993.

\bibitem{Murata96}
T.~{Murata} and H.~{Ishibuchi}.
\newblock Positive and negative combination effects of crossover and mutation
  operators in sequencing problems.
\newblock In {\em Proc. of Congress on Evolutionary Computation (CEC'96)},
  pages 170--175, May 1996.

\bibitem{caceres2017experimental}
L.~{P{\'e}rez C{\'a}ceres}, M.~L{\'o}pez-Ib{\'a}{\~n}ez, H.~Hoos, and
  T.~St{\"u}tzle.
\newblock An experimental study of adaptive capping in irace.
\newblock In {\em Proc. of International Conference on Learning and Intelligent
  Optimization (LION'17)}, pages 235--250. Springer, 2017.

\bibitem{pushak2018algorithm}
Y.~Pushak and H.~Hoos.
\newblock Algorithm configuration landscapes.
\newblock In {\em Prof. of Parallel Problem Solving from Nature (PPSN'18)},
  pages 271--283. Springer, 2018.

\bibitem{rechenberg1989evolution}
I.~Rechenberg.
\newblock Evolution strategy: Nature's way of optimization.
\newblock In {\em Optimization: Methods and applications, possibilities and
  limitations}, pages 106--126. Springer, 1989.

\bibitem{ShahriariSWAF16}
B.~Shahriari, K.~Swersky, Z.~Wang, R.~P. Adams, and N.~de~Freitas.
\newblock {Taking the Human Out of the Loop: {A} review of Bayesian
  optimization}.
\newblock {\em Proc. of the {IEEE}}, 104(1):148--175, 2016.

\bibitem{spears1993crossover}
W.~M. Spears.
\newblock Crossover or mutation?
\newblock In {\em Proc. of Foundations of genetic algorithms (FOGA'93)}, pages
  221--237. Elsevier, 1993.

\bibitem{Stutzle2019}
T.~St{\"u}tzle and M.~L{\'o}pez-Ib{\'a}{\~n}ez.
\newblock {\em Automated Design of Metaheuristic Algorithms}, pages 541--579.
\newblock Springer, Cham, 2019.

\bibitem{sudholt2017crossover}
D.~Sudholt.
\newblock How crossover speeds up building block assembly in genetic
  algorithms.
\newblock {\em Evolutionary computation}, 25(2):237--274, 2017.

\bibitem{Sudholt2020chapterdiversity}
D.~Sudholt.
\newblock The benefits of population diversity in evolutionary algorithms: A
  survey of rigorous runtime analyses.
\newblock In B.~Doerr and F.~Neumann, editors, {\em Theory of Evolutionary
  Computation: Recent Developments in Discrete Optimization}, pages 359--404.
  Springer, 2020.

\bibitem{tanabe2020analyzing}
R.~Tanabe.
\newblock Analyzing adaptive parameter landscapes in parameter adaptation
  methods for differential evolution.
\newblock In {\em Proc. of the Genetic and Evolutionary Computation Conference
  (GECCO 20')}, pages 645--653, 2020.

\bibitem{thierens2011optimal}
D.~Thierens and P.~A. Bosman.
\newblock Optimal mixing evolutionary algorithms.
\newblock In {\em Proc. of Genetic and Evolutionary Computation Conference
  (GECCO'11)}, pages 617--624, 2011.

\bibitem{TinosWCO21}
R.~Tin{\'{o}}s, D.~Whitley, F.~Chicano, and G.~Ochoa.
\newblock Partition crossover for continuous optimization: {ePX}.
\newblock In {\em Proc. of Genetic and Evolutionary Computation Conference
  (GECCO'21)}, pages 627--635. ACM, 2021.

\bibitem{TintosWC15}
R.~Tin{\'{o}}s, L.~D. Whitley, and F.~Chicano.
\newblock Partition crossover for pseudo-boolean optimization.
\newblock In {\em Proc. of Foundations of Genetic Algorithms (FOGA'15)}, pages
  137--149. ACM, 2015.

\bibitem{van2019automatic}
B.~van Stein, H.~Wang, and T.~B{\"a}ck.
\newblock Automatic configuration of deep neural networks with parallel
  efficient global optimization.
\newblock In {\em Proc. of International Joint Conference on Neural Networks
  (IJCNN'19)}, pages 1--7. IEEE, 2019.

\bibitem{WangEB18}
H.~Wang, M.~Emmerich, and T.~B{\"{a}}ck.
\newblock {Cooling Strategies for the Moment-Generating Function in Bayesian
  Global Optimization}.
\newblock In {\em Proc. of Congress on Evolutionary Computation (CEC'18)},
  pages 1--8. {IEEE}, 2018.

\bibitem{wang2017new}
H.~Wang, B.~van Stein, M.~Emmerich, and T.~B\"ack.
\newblock A new acquisition function for {Bayesian} optimization based on the
  moment-generating function.
\newblock In {\em Proc. of International Conference on Systems, Man, and
  Cybernetics (SMC'17)}, pages 507--512. IEEE, 2017.

\bibitem{IOHanalyzer}
H.~Wang, D.~Vermetten, F.~Ye, C.~Doerr, and T.~B{\"{a}}ck.
\newblock {IOHanalyzer}: Performance analysis for iterative optimization
  heuristic.
\newblock {\em ACM Transactions on Evolutionary Learning and Optimization},
  2022.

\bibitem{WmodelInstancesASoC}
T.~Weise, Y.~Chen, X.~Li, and Z.~Wu.
\newblock Selecting a diverse set of benchmark instances from a tunable model
  problem for black-box discrete optimization algorithms.
\newblock {\em Applied Soft Computing}, 92:106269, 2020.

\bibitem{weise2018difficult}
T.~Weise and Z.~Wu.
\newblock Difficult features of combinatorial optimization problems and the
  tunable w-model benchmark problem for simulating them.
\newblock In {\em Proc. of Genetic and Evolutionary Computation Conference
  (GECCO'18, Companion Material)}, pages 1769--1776. ACM, 2018.

\bibitem{ye2019interpolating}
F.~Ye, C.~Doerr, and T.~B{\"a}ck.
\newblock Interpolating local and global search by controlling the variance of
  standard bit mutation.
\newblock In {\em Proc. of IEEE Congress on Evolutionary Computation (CEC'19)},
  pages 2292--2299. IEEE, 2019.

\bibitem{Dataref}
F.~Ye, C.~Doerr, H.~Wang, and T.~B\"ack.
\newblock {Data sets for the study "Automated Configuration of Genetic
  Algorithms by Tuning for Anytime Performance"}.
\newblock \url{https://doi.org/10.5281/zenodo.4823492}, May 2021.

\bibitem{Yeppsn2020}
F.~Ye, H.~Wang, C.~Doerr, and T.~B{\"a}ck.
\newblock Benchmarking a $(\mu +\lambda)$ genetic algorithm with configurable
  crossover probability.
\newblock In {\em Proc. of Parallel Problem Solving from Nature (PPSN'20)},
  pages 699--713. Springer, 2020.

\end{thebibliography}

\begin{IEEEbiography}[{\includegraphics[width=1in,height=1.25in,clip,keepaspectratio]{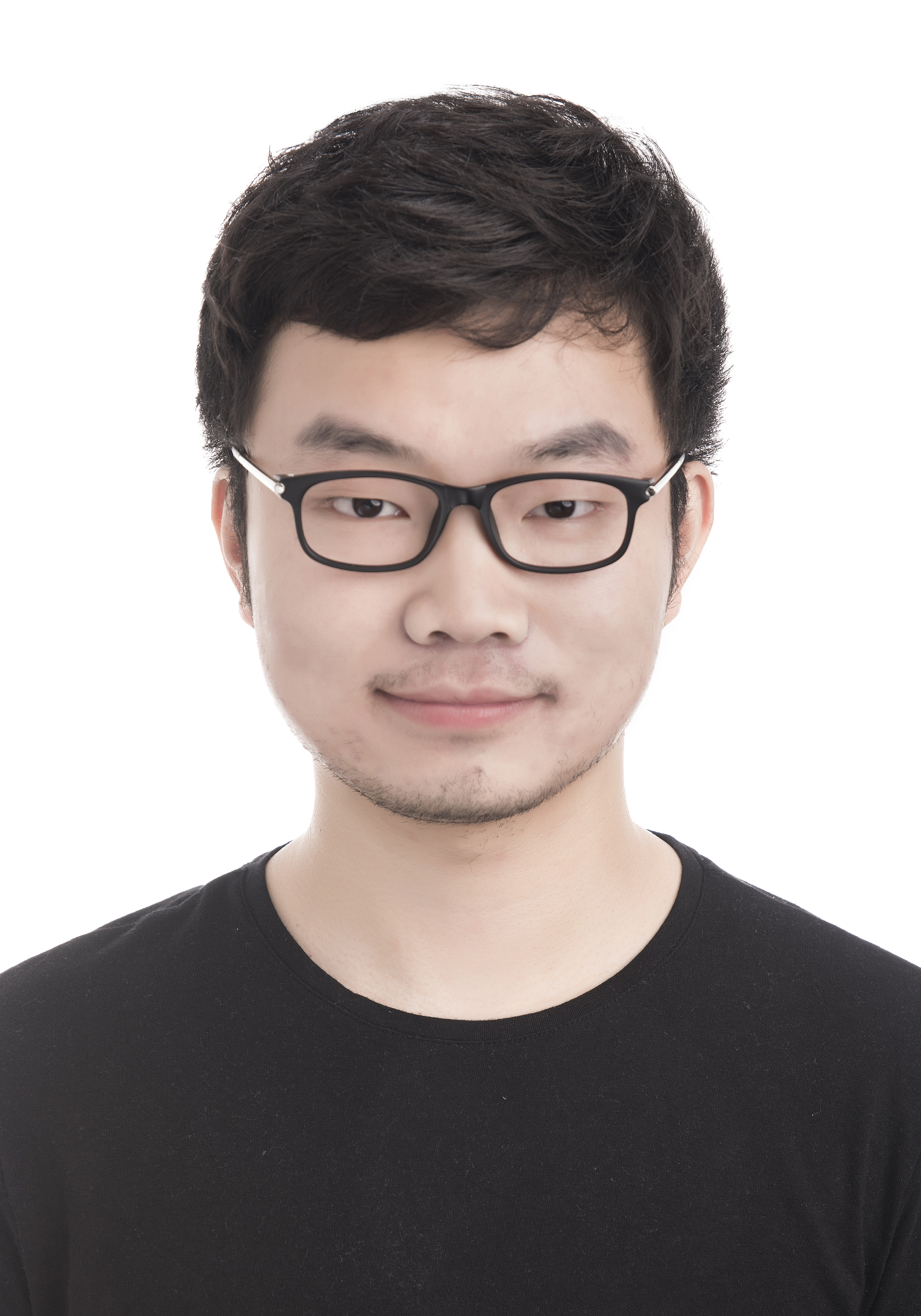}}]{Furong Ye,} 
is a PostDoc at the Leiden Institute of Advanced Computer Science (LIACS), after finishing his PhD study at LIACS. His PhD topic is ``Benchmarking discrete optimization heuristics: From building a sound experimental environment to algorithm configuration''. He is part of the core development team of IOHprofiler, with a focus on the IOHexperimenter. His research interests are the empirical analysis of algorithm performance and (dynamic) algorithm configuration.
\end{IEEEbiography}

\begin{IEEEbiography}[{\includegraphics[width=1in,height=1.25in,clip,keepaspectratio]{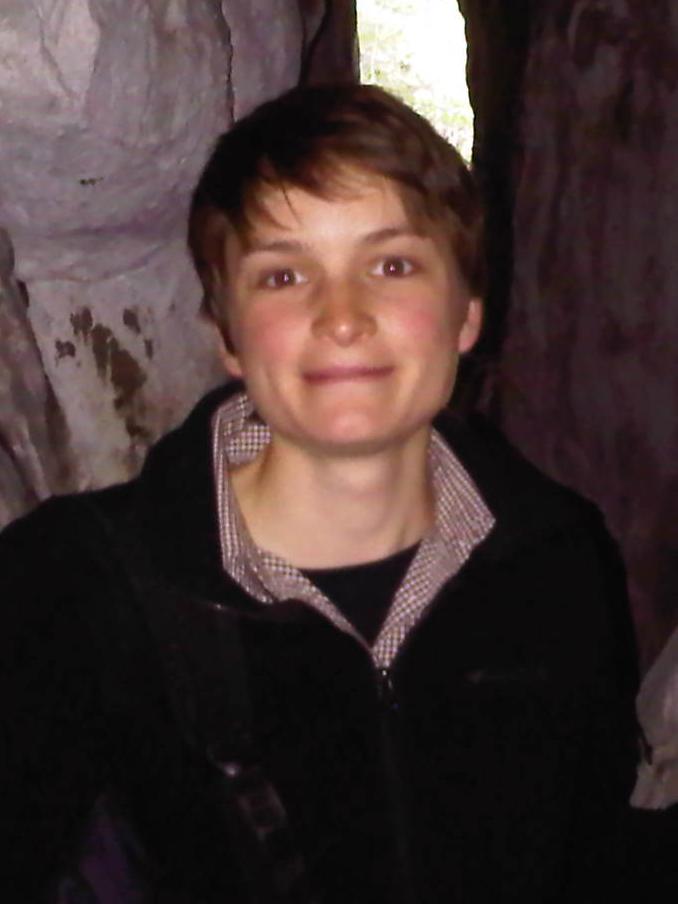}}]{Carola Doerr,} 
formerly Winzen, is since 2013 a permanent CNRS researcher at Sorbonne Universit\'e in Paris, France. Carola's main research activities are in the mathematical analysis of randomized algorithms, with a strong focus on evolutionary algorithms and other sampling-based optimization heuristics. She is also interested in all aspects of benchmarking these algorithms. Carola obtained her PhD in Computer Science from Saarland University and the Max Planck Institute for Informatics in 2011 and she successfully defended her habilitation (HDR) at Sorbonne Universit\'e in 2020.  

She is associate editor of IEEE Transactions on Evolutionary Computation, of ACM Transactions on Evolutionary Learning and Optimization, editorial board member of the Evolutionary Computation journal, advisory board member of the Springer Natural Computing Book Series, and
guest editor of a special issue in TEVC and two special issues in Algorithmica. She is a founding and coordinating member of the \href{https://sites.google.com/view/benchmarking-network/}{Benchmarking Network}, an initiative created to consolidate and to stimulate activities on benchmarking sampling-based optimization heuristics. 
\end{IEEEbiography}

\begin{IEEEbiography}[{\includegraphics[width=1in,height=1.25in,clip,keepaspectratio]{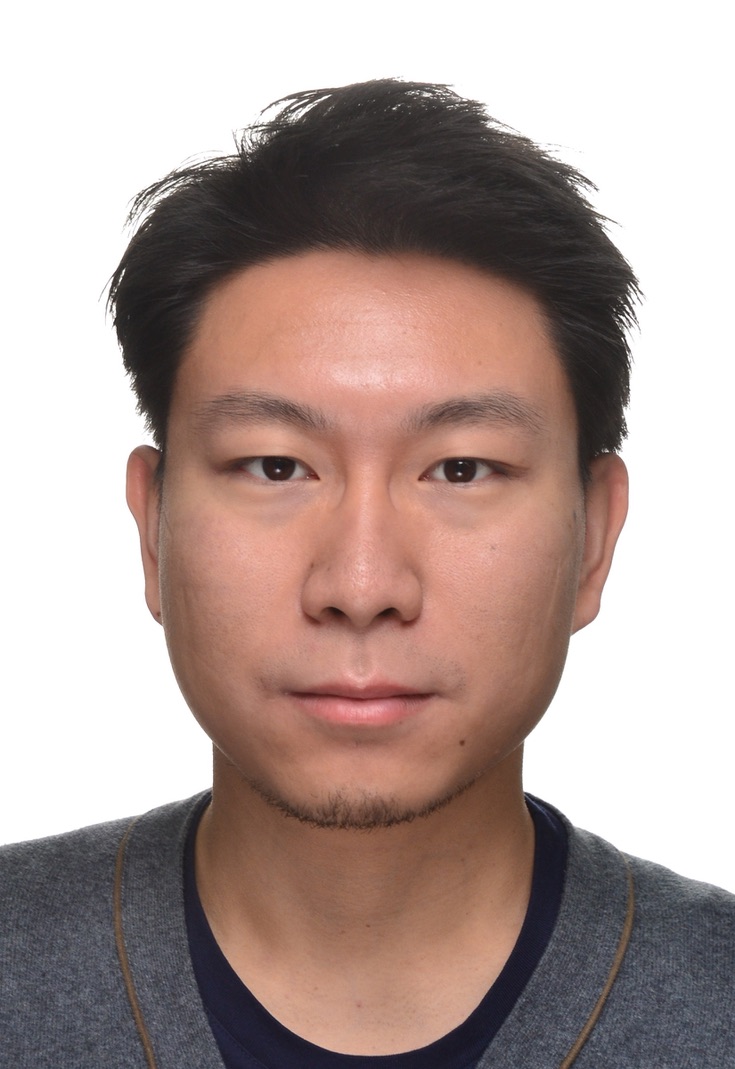}}]{Hao Wang} is an assistant professor of computer science in Leiden University since September 2020, focusing black-box optimization, Bayesian optimization, multi-objective optimization, and variational quantum algorithms. Hao obtained his Ph.D. (cum laude) at Leiden University in 2018, followed by two postdoctoral appointments: at Leiden University (2018.05 – 2019.12) and LIP6, Sorbonne Universit\'e, Paris, France (2020.01 – 2020.08). He served as the proceedings chair for the PPSN 2020 conference and will be organizing the EMO (Evolutionary Multi-Objective Optimization) 2023 international conference as one of the general co-chairs. He received the best paper award at the PPSN (Parallel Problem Solving from Nature) 2016 conference for proposing new measures to understand the difficulties of multi-objective optimization problems. 
\end{IEEEbiography}

\begin{IEEEbiography}[{\includegraphics[width=1in,height=1.25in,clip,keepaspectratio]{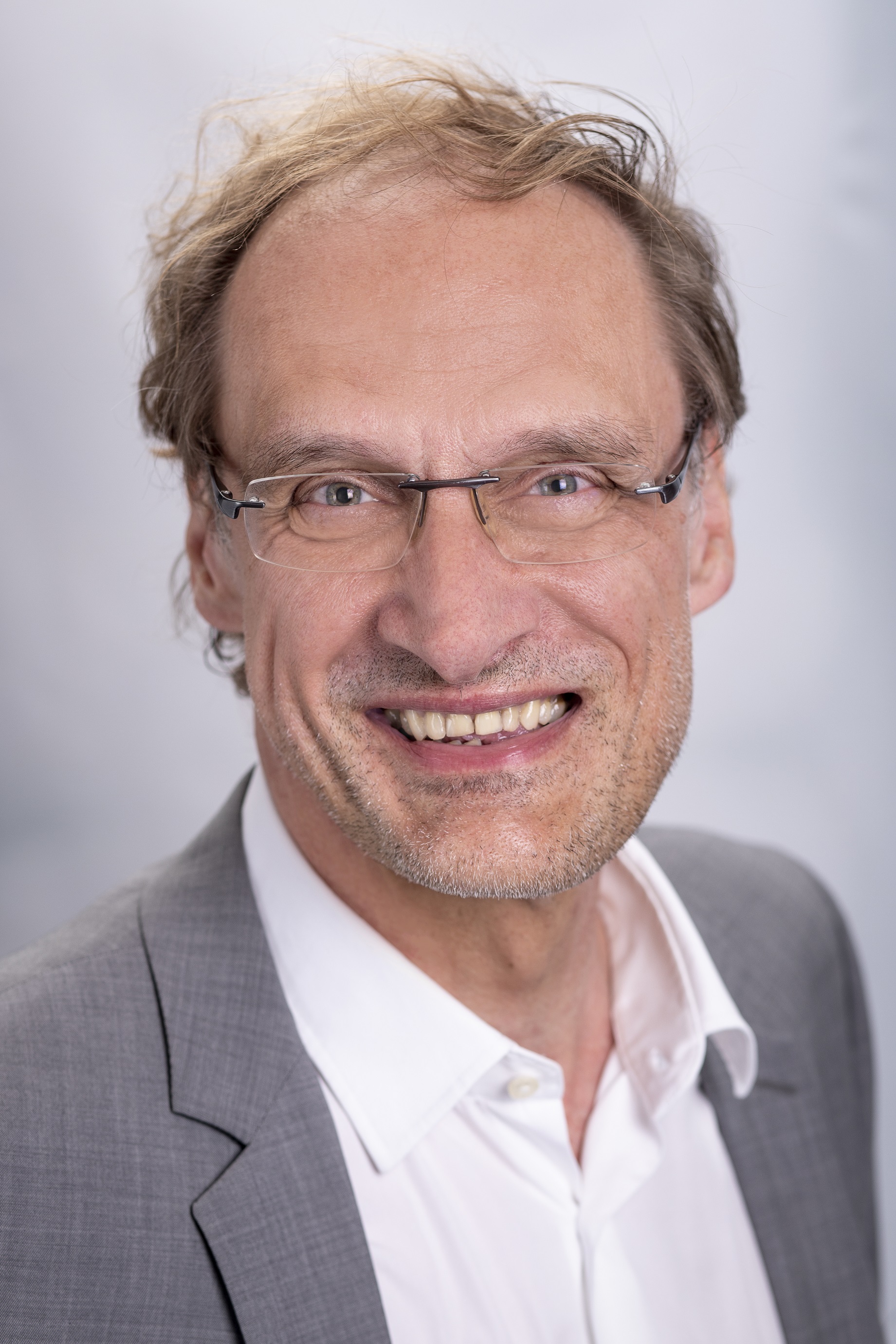}}]{Thomas B\"ack,}
 is Full Professor of Computer Science at the Leiden Institute of Advanced Computer Science (LIACS), Leiden University, The Netherlands, since 2002. 
 He received the Diploma degree in computer science in 1990 and the PhD~degree in 1994, both from the University of Dortmund, Germany. 
 His research interests include evolutionary computation, machine learning, and their real-world applications, especially in sustainable smart industry and health.
 
 In 2021, Thomas has been elected as member of the Royal Netherlands Academy of Aarts and Sciences (KNAW). 
 He was a recipient of the IEEE Computational Intelligence Society (CIS) Evolutionary Computation Pioneer Award in 2015, was elected as Fellow of the International Society of Genetic and Evolutionary Computation in 2003, and received the best PhD~thesis award from the German society of Computer Science (Gesellschaft f\"ur Informatik, GI) in 1995. 
 
 He currently serves as an Associate Editor for the IEEE Transactions on Evolutionary Computation and Artificial Intelligence Review journals, and as area editor for the ACM Transactions on Evolutionary Learning and Optimization. He is co-editor of the Natural Computing Book Series, Handbook of Evolutionary Computation, Handbook of Natural Computing, and author of Evolutionary Computation in Theory and Practice. 
\end{IEEEbiography}

\clearpage 
\section*{Figure Captions and Table Captions}
\noindent \textbf{Fig. 1.}
Fixed-target ERT values of the configurations suggested for F10. The suffix ``-ERT/AUC'' indicates which cost metric was used during the tuning.\\

\noindent \textbf{Fig. 2.}
Relative ERT values of the GAs obtained by Irace, MIES, and MIP-EGO using ERT as the cost metric compared to the ERT of the GAs obtained by the same method when using AUC as the cost metric during the configuration process. Plotted values are (ERT$_{\text{using ERT}}$-ERT$_{\text{using AUC}}$)/ERT$_{\text{using AUC}}$, capped at $-1$ and $1$. Positive values therefore indicate that configurator obtains better results when configuring AUC.\\

\noindent \textbf{Fig. 3.}
The relative deviation from the best-known ERT value of the GAs obtained during the configuration process of Irace for tuning the \mlga for \onemax in dimension $n=100$, with the objective to minimize the ERT for the optimum $f(x)=100$. The maximal number of configurations that can be tested by Irace is set to 5\,000. The figure is produced by the acviz tool, and illustrations for the details of plot representation can be found in Sec. 3 of~\cite{SouzaEtAl2020acviz}.\\

\noindent \textbf{Fig. 4.}
Violin plots of first hitting times for the configurations found by Irace when tuning for ERT and AUC, respectively. Only showing results for problems on which Irace-ERT outperforms Irace-AUC. 
Results are from the 100 independent validation runs. Targets are listed in Tab.~\ref{tab:targets-f}, and the configurations of the GAs can be found in Tab.~\ref{tab:config}. For each run, values are capped at the budget $50\,000$ if the algorithm cannot find the target.\\

\noindent \textbf{Fig. 5.} 
Violin plots of first hitting times for the configurations found by Irace when tuning for ERT and AUC, respectively, for problems on which Irace-AUC outperforms Irace-ERT. 
Results are from the 100 independent validation runs. Targets are listed in Tab.~\ref{tab:targets-f}, and the configurations of the GAs can be found in Tab.~\ref{tab:config}. For each run, values are capped at the budget $50\,000$ if the algorithm cannot find the target.\\
    
\noindent \textbf{Fig. 6.}
Fixed-target ERT values of the GAs listed in Tab.~\ref{tab:config} for F21.\\

\noindent \textbf{Fig. 7.}
Fixed-target ERT values of the GAs listed in Tab.~\ref{tab:config} for F8.\\

\noindent \textbf{Fig. 8.}
Fixed-target ERT values of the GAs listed in Tab.~\ref{tab:config} for F7.\\

\noindent \textbf{Fig. 9.}
ERT values ($y$-axis) of the GAs obtained by Irace for \onemax and \leadingones in dimension $n=100$, for different cutoff time $B$ that the GAs can spend to find the optimum ($x$-axis). Showing results for $B \in \{(0.5 + 0.1t) \text{ERT}_{(1+1) \text{ EA}} \mid  t \in [0..15]\}$. For comparison, the ERT values of the \oea are plotted by horizontal red lines. Results are for the best found configurations obtained from $r=20$ independent runs of Irace, and each of the ERT values is with respect to 100 independent validation runs.\\

\noindent \textbf{Fig. 10.}
ERT values ($y$-axis) of the GAs obtained by Irace with different configuration budgets $B_T$ (the number of configurations that Irace can test, $x$-axis). Results are for $B_T \in \{(0.5 + 0.25t) 5\,000 \mid t \in [0..4]\}$. Each ERT value is for the 100 validation runs of the configuration suggested by Irace after a single run, i.e., one for each budget.\\

\noindent\textbf{Table. 1.}
Target values used to compute the ERT metric on each problem.\\

\noindent \textbf{Table. 2.}
Configurations of the \mlga obtained by grid search, Irace, MIP-EGO, and MIES. C indicates the cost metric used by the configurators. Results for maximizing AUC are  obtained independently from those obtained for minimizing ERT.\\

\noindent \textbf{Table. 3.}
Absolute ERT and AUC values for the \oea and relative improvement of ERT and AUC for the configurations suggested by the four configuration methods, in comparison against the \oea values. Compared to the ERT and AUC values of the \oea on each problem (indicated by ``EA''), the relative improvement obtained from the automated configuration are shown for each AC method, where both measures are calculated from hitting times of $100$ validation runs for \oea and AC methods. We also indicate the statistical significance in the empirical distributions of the hitting time ($^{****}$ for $p<0.001$, $^{***}$ for $p<0.01$, $^{**}$ for $<0.01$, and $^{*}$ for $p<0.05$) for each pair of the AC result and that of the \oea. The Mann–Whitney U test is applied with the Benjamini and Hochberg method for all 120 pairwise comparisons to control the false discovery rate. Runs are cut off at $50\,000$ evaluations if it does not hit the final target. The significant comparisons are colour-coded with respect to the relative improvement, where a darker colour signifies a more considerable improvement.
\end{document}